\pdfoutput=1

\documentclass[11pt]{article}

\usepackage[preprint]{acl}

\usepackage{times}
\usepackage{latexsym}
\usepackage[T1]{fontenc}
\usepackage[utf8]{inputenc}
\usepackage{microtype}
\usepackage{inconsolata}
\usepackage{graphicx}
\usepackage{xcolor}

\graphicspath{ {./Figures/} }

\usepackage{fontawesome5}

\title{Learning vs Retrieval: The Role of In-Context Examples in Regression with Large Language Models}

\author{
 \textbf{Aliakbar Nafar\textsuperscript{1}},
 \textbf{Kristen Brent Venable\textsuperscript{2,3}},
 \textbf{Parisa Kordjamshidi\textsuperscript{1}}\\
 \textsuperscript{1}Michigan State University\\
 \textsuperscript{2}Florida Institute for Human and Machine Cognition\\
 \textsuperscript{3}University of West Florida
\\
\\
\{nafarali, kordjams\}@msu.edu, bvenable@ihmc.org
}

\begin{document}
\maketitle
\begin{abstract}

Generative Large Language Models (LLMs) are capable of being in-context learners. However, the underlying mechanism of in-context learning (ICL) is still a major research question, and experimental research results about how models exploit ICL are not always consistent. In this work, we propose a framework for evaluating in-context learning mechanisms, which we claim are a combination of retrieving internal knowledge and learning from in-context examples by focusing on regression tasks. First, we show that LLMs can solve real-world regression problems and then design experiments to measure the extent to which the LLM retrieves its internal knowledge versus learning from in-context examples. We argue that this process lies on a spectrum between these two extremes. We provide an in-depth analysis of the degrees to which these mechanisms are triggered depending on various factors, such as prior knowledge about the tasks and the type and richness of the information provided by the in-context examples. We employ three LLMs and utilize multiple datasets to corroborate the robustness of our findings. Our results shed light on how to engineer prompts to leverage meta-learning from in-context examples and foster knowledge retrieval depending on the problem being addressed.
\end{abstract}


\section{Introduction}

The emergence of transformers~\cite{NIPS2017_3f5ee243} has revolutionized natural language processing, leading to the development of LLMs such as GPTs~\cite{brown2020languagemodelsfewshotlearners}. In addition to their impressive zero-shot performance, these models demonstrated the capability of in-context learning (ICL), by which they learn a task from examples provided in the context of the prompt~\cite{brown2020languagemodelsfewshotlearners}. In tasks where both zero-shot and ICL settings are applicable, ICL consistently outperforms the zero-shot setting~\cite{brown2020languagemodelsfewshotlearners,liu-etal-2022-makes}. However, the inner mechanism of ICL in LLMs remains an open area of research. 

Broadly speaking, current research identifies two main approaches to explain the ICL mechanism~\cite{dong-etal-2024-survey,li-etal-2024-language}: 1) Meta-learning, 2) Knowledge retrieval. With meta-learning~\cite{schmidhuber1987evolutionary}, the LLM \textbf{learns} a new \textbf{unseen} pattern from the in-context examples, allowing it to generalize to new inputs that follow the same pattern. In contrast, with knowledge retrieval, the LLM uses the in-context examples as cues to access and apply information already stored within its pre-trained parameters without inferring new patterns from the provided examples. For instance, given examples of input/output number pairs in the context, meta-learning allows the model to learn their relationship and apply it to new inputs. However, if the prompt informs the model that the pairs are the earth's population given a year, the LLM utilizes its knowledge retrieval and infers that the function is exponential and plateaus at around 8-10 billion after the year 2000.

Contrary to current research, we propose a different approach, arguing that ICL is not merely learning or retrieving knowledge, but rather, its behavior lies on a \textit{spectrum} between the two, which can be adjusted depending on various factors. Before exploring our proposed approach, we elaborate on the hypotheses mentioned above.~\footnote{See Appendix~\ref{appendix:related_work} for a detailed discussion on related work.}

\begin{figure*}[!ht]
    \centering
    \includegraphics[width=0.9\textwidth]{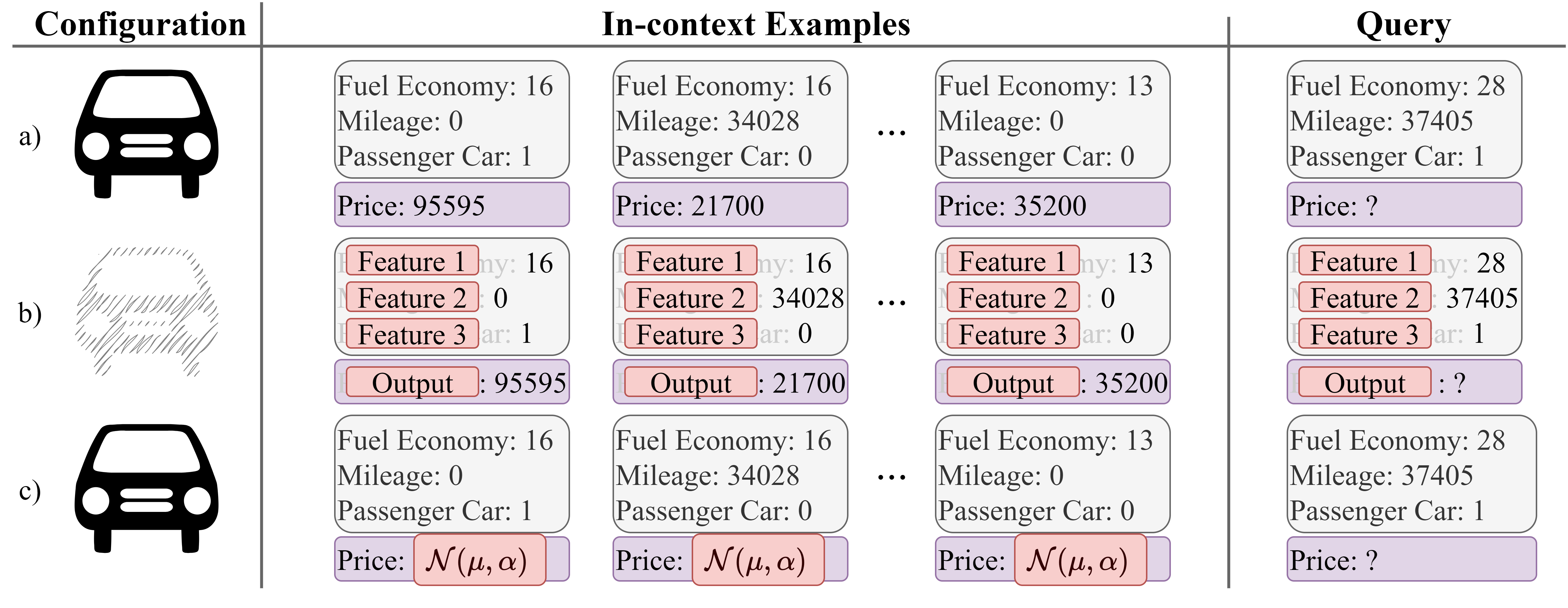}
    \caption{The three main prompt configurations: In configuration a) the actual names of the features and the output are known, and the LLM is asked to guess the ``price of a used Toyota or Maserati in 2019''. Configuration b) is similar to a) except that the feature names are anonymized. Here, the LLM is asked to estimate the ``Output''. In Configuration c), we replace the real prices of in-context examples with randomly (Gaussian) generated numbers.} 
    \label{fig:intro}
\end{figure*}

The first hypothesis suggests that transformers are effective meta-learners, and LLMs can generate accurate predictions based solely on the given input-output pairs. \cite{bai2023transformers} theoretically proves and practically tests the capability of an encoder-based transformer to implement generalized linear models in-context. \cite{Garg2022WhatCT} uses a decoder-based model, a GPT-2 architecture modified for regression, that uses the input features inside the embeddings (instead of using tokens) and outputs a number. Their model's performance surpasses a 2-layer Multi-Layer Perceptron (MLP) and a decision tree by meta-learning. Further, \cite{vacareanu2024words} directly uses LLMs to test regression capabilities using a limited number of features (less than 3) and concludes that LLMs are capable regressors based on in-context examples. However, according to our findings, their claim of avoiding data contamination is not well-supported. We note that none of these research works use realistic datasets or consider the combination of meta-learning with knowledge retrieval.

The second approach emphasizes knowledge retrieval while downplaying the learning aspect. For example,~\cite{min-etal-2022-rethinking} examines 12 LLMs across various classification tasks, concluding that altering output labels, which are needed for learning, has no impact on performance. They argue that the crucial elements of ICL are limited to defining the label space, input distribution, and overall task format. Meanwhile, \cite{kossen2024incontextlearninglearnslabel} argues that LLMs utilize the labels but in an arbitrary manner when experimenting with the same classification tasks. \cite{li-etal-2024-language} further dismisses the significance of output labels and instead proposes two knowledge retrieval approaches. These include the retrieval of pre-existing solutions from training data~\cite{min-etal-2022-rethinking,wang2024largelanguagemodelslatent,wies2023learnabilityincontextlearning}, and a novel \textit{solution composition} approach, inspired by ~\cite{hahn2023theoryemergentincontextlearning}, which proposes that learned solutions can be combined. While this work offers valuable insights into ICL mechanisms, its findings are limited to a narrow selection of datasets and models. Here, the exclusive use of LLaMA 2~\cite{touvron2023LLaMA}, which is known to struggle with long token contexts~\cite{machlab2024llmincontextrecallprompt,meta_LLaMA_issue_887} results in some different findings compared to our work. For instance, they report that remapping the inputs of in-context input-output pairs to alternative text reduces the accuracy to that of a random model (even with up to 40 in-context examples). On the other hand, under similar conditions, our experiments indicate that in newer models, learning occurs with optimal performance. We refrain from making specific claims about task selection or composition in our work. Instead, we categorize these processes under the broader umbrella of \textit{knowledge retrieval}~\footnote{See Appendix~\ref{appendix:aclpaper} for a more detailed analysis of the discrepancies between our findings and those of ~\cite{li-etal-2024-language}.}. Similarly, to investigate ICL, \cite{pan-etal-2023-context} uses simple classification tasks and older models (e.g., GPT-3 Ada), and as a result, its findings are partially different from ours.

We propose a different hypothesis that resolves the contradictions in the research community and support our claims with extensive empirical testing. We argue that ICL is not merely learning or retrieving knowledge but uses a combination of the two, which lies on a spectrum determined by factors we can manipulate. In this regard, we propose an evaluation framework and conduct a comparative study of different LLMs and datasets, focusing specifically on regression problems as our testing ground. In our evaluation framework, we query the LLM to estimate an output number based on a set of (feature, value) pairs given a set of ICL examples, as shown in Figure~\ref{fig:intro}(a). 

We opted for regression tasks for the following reasons: This choice aligns our work with the majority of related ICL meta-learning research, thus facilitating direct comparisons and building upon existing findings. These include works that alter the output module of LLMs, such as \cite{Garg2022WhatCT} and those that use the token generation output of LLMs as is, such as \cite{vacareanu2024words}. Moreover, \cite{Bhattamishra2023UnderstandingIL} demonstrates that the ICL capability of LLMs is independent of their input/output modules and is present in the intermediary layers of the LLMs. Finally, while LLMs are capable of regression (even when inputs are given to them in a textual format~\cite{vacareanu2024words}), this task still has a complex output space (e.g., continuous or unbounded), which is challenging for LLMs~\cite{Fang2024}.   

We show that LLMs can perform regression on realistic datasets and measure the extent to which the LLM retrieves its internal knowledge versus learning from in-context examples. We provide an in-depth analysis of the degrees to which these mechanisms are triggered depending on the factors we use in our framework: the number of (feature, value) pairs, the number of in-context examples, and the prompting strategies.


In summary, our contributions are as follows: \textbf{1)} We demonstrate that LLMs can effectively learn from regression examples of \textbf{realistic} datasets in-context, extending previous work on synthetic data to more practical scenarios.
\textbf{2)} We propose a new hypothesis about the ICL mechanism that combines both learning and knowledge retrieval during the LLM inference, reconciling the research community's findings.
\textbf{3)} We introduce an evaluation framework that allows for systematic comparison of ICL mechanisms across different LLMs, datasets, and prompt configurations\footnote{The code and analytical tools are available at \href{https://github.com/HLR/LvsR-LLM/}{\faGithub}.}.
\textbf{4)} We provide a comprehensive analysis of how LLMs balance internal knowledge retrieval and learning from in-context examples and propose prompt engineering techniques to control them.



\section{Problem Setting}

Our study focuses on a regression task where we use LLMs to predict numerical outputs based on the given inputs. We assume  a regression dataset, $D = \{(x_1, y_1), (x_2, y_2), \ldots, (x_n, y_n)\}$, comprising input-output pairs, is given. Each $x_i$ comprises a set of pairs $(f_{ij},v_{ij})$ where $f_{ij}$ represents the feature name and $v_{ij}$ is the numerical value of $f_{ij}$. The target variable $y_i$ is the numerical value of the output. For example, if we consider a dataset providing the price of used cars given their fuel economy and mileage, we could have an input with (feature, value) pairs, (Fuel Economy,$16$) and (Mileage, $0$), and an output, $95595$, indicating the price. 

To resemble a regression problem setting, we present the LLM with a subset of $m$ input-output examples $\{(x_1, y_1), (x_2, y_2), \ldots, (x_m, y_m)\}$ taken from the in-context split of $D$ for learning. Then,  we query the LLM with the feature pairs of $x$ taken from the test split of $D$ to obtain the answer $y$ as the output.\footnote{We note that since we do not change the architecture of the LLMs, all numbers must be represented to and read by the LLM in textual form.} Figure~\ref{fig:intro}(a) depicts our problem setting in the context of a regression task that involves predicting the ``price of a used Toyota or Maserati in 2019'', which is the target variable name, given three (feature, value) pairs with a set of provided in-context examples. When $m$ equals 0, we obtain the zero-shot setting without prior examples.

\subsection{Prompt Configurations}
We employ three main prompt configurations for prompting the models, as shown in Figure~\ref{fig:intro}. These configurations vary in different ways, such as hiding the real name of features (Figure~\ref{fig:intro}(b)) or the actual ground truth (Figure~\ref{fig:intro}(c)). We also prompt the LLM with simple numerical generation as a baseline in our experiments. This results in a total of four prompt configurations. We also consider other prompt configurations which either served as ablations or didn't provide any significant insights. In each prompt configuration, the context of the prompt comprises a task instruction, in-context examples (sample regression task input-outputs), and a query. The task instruction asks the LLM to estimate the objective of the dataset (such as the price of a used car) based on the given features by \textbf{just providing a number and no explanation as the output}. Each of these parts can be changed or removed according to the prompt configuration.

\noindent\textbf{Named Features} (Configuration a): This is our most straightforward prompt configuration. It reveals the actual names of the features and the required target variable (depending on the dataset) in the prompt. For instance, in the case of a dataset about the price of used cars, the LLM is instructed to estimate a ``Used Car Price'' based on ``City fuel Economy'', ``Mileage'' and ``Passenger Car Classification''. Following this instruction, sample input-output examples are given to the model, and then the query is asked as shown in Figure~\ref{fig:intro}(a).

\noindent\textbf{Anonymized Features} (Configuration b): In this setup, which is shown in Figure~\ref{fig:intro}(b), we present the LLM with input-output pairs where the actual name of the features and the target variable are changed to  ``Feature \#'' and ``Output'', respectively. The instruction asks the LLM to estimate the ``Output'' based on the given ``Features''. Here, the LLM can not use its domain knowledge, and \textbf{can only} use the provided number of the features.

\noindent\textbf{Randomized Ground Truth} (Configuration c): In our final main prompt configuration shown in Figure~\ref{fig:intro}(c), we maintain the named features but replace the ground truth values with randomly generated numbers (independent of the feature values). These random numbers are generated using a Gaussian distribution based on the dataset statistics. This setup mainly serves as a control configuration to test to what degree the LLMs are truly learning from the provided ground truth in the examples.

\noindent\textbf{Direct Question Answering} (Direct QA): To establish an LLM baseline, we ask LLMs to estimate the target variable based on the given named features without any in-context examples ($m=0$). We also define the scope of the output by providing the mean and the standard deviation of the questioned dataset in the instruction to the LLM. An example of this added information is: ``Estimate the insurance cost of this person given the information. An issuance cost is typically around $13270.42$ with a standard deviation of $12110.01$''.

\subsection{Models and Metrics}

We evaluate several LLMs, including LLaMA 3 70B~\cite{meta2024LLaMA3}, GPT3.5~\cite{brown2020languagemodelsfewshotlearners}, and GPT4~\cite{openai2023gpt4}. We initially considered smaller LLMs such as Mistral 7B~\cite{jiang2023mistral}, but found their performance for regression tasks to be insufficient and consequently excluded them. To provide a comprehensive comparison, we also employed classical machine learning techniques alongside these LLMs. Specifically, we utilized a straightforward Ridge regression~\cite{hoerl1970ridge} and the more advanced RandomForest model~\cite{Breiman2001}. Further details about these models, such as their hyper-parameters, are included in Appendix~\ref{appendix:Hyper_Parameters}.

In our analysis, for each dataset, we test the LLMs with a mix of \textbf{factors} (we call these factors to distinguish them from the features $f_{ij}$ defined in our problem setting). The first factor is the \emph{prompt configurations}, which can be Named Feature, Anonymized Features, Randomized Ground Truth, or Direct QA. The second and third factors are the \emph{number of in-context examples} and the \emph{number of (feature, value) pairs}, which we will refer to as the number of features. The number of in-context examples is 0, 10, 30, and 100, and the number of features is 1, 2, and 3 in our experiments. These feature numbers indicate using the first feature (F1), the first two features (F2), and all three features (F3) (features are ordered and sorted in decreasing order of
importance). Not all these factors can be used together as the 0 in-context examples factor can only be used with the Direct QA prompt configuration, and other prompt configurations can not be used with the 0 in-context examples factor. To assess the performance of the LLMs and the machine learning models on regression tasks, we use Mean Squared Error (MSE) as our main evaluation metric. We also calculate and report the coefficient of determination $R^2$ and Mean Absolute Error (MAE) in Appendix~\ref{appendix:other_metrics}.

\section{Experiments}

This section presents our experimental findings, highlighting the interplay between various factors that influence the performance of LLMs across regression tasks and datasets.

\subsection{Datasets}
\label{sec:datasets}

To assess the LLMs' performance across diverse domains and complexity levels, we select three regression datasets and three of their most important features, where feature importance was calculated with RandomForest~\cite{louppe2013understanding}: 1) Admission Chance: the \textit{likelihood of admission to graduate programs for Indian students}~\cite{chanceofadmission8862140} with three highly correlated features. Since this dataset is about Indian students, there is a lower chance of being observed in the LLM's training data, reflecting the general imbalance that skews towards USA-centric data~\cite{zhu2023extrapolatinglargelanguagemodels}. 2) Insurance Cost: focuses on predicting a similar distribution of the \textit{annual individual medical costs billed by health insurance companies in the USA}, drawing from demographic statistics compiled by the U.S. Census Bureau~\cite{lantz2013machine}. For this dataset, the first feature is the most important one by far. 3) Used Car Prices: the estimation of the \textit{price of a used Toyota or Maserati car in 2019}. Here, the first two features are the most impactful ones. \footnote{All datasets used in this study are publicly available. For references and details of our pre-processing code, visit \href{https://github.com/HLR/LvsR-LLM/tree/main/Datasets}{\faGithub}.} These datasets are pre-processed to select their most important features. We divide each dataset into two splits: an in-context subset comprising 100 instances and a test subset with 300 instances. See Appendix~\ref{appendix:datasets} for details.

\begin{figure}[h]
    \centering
    \includegraphics[width=0.47\textwidth]{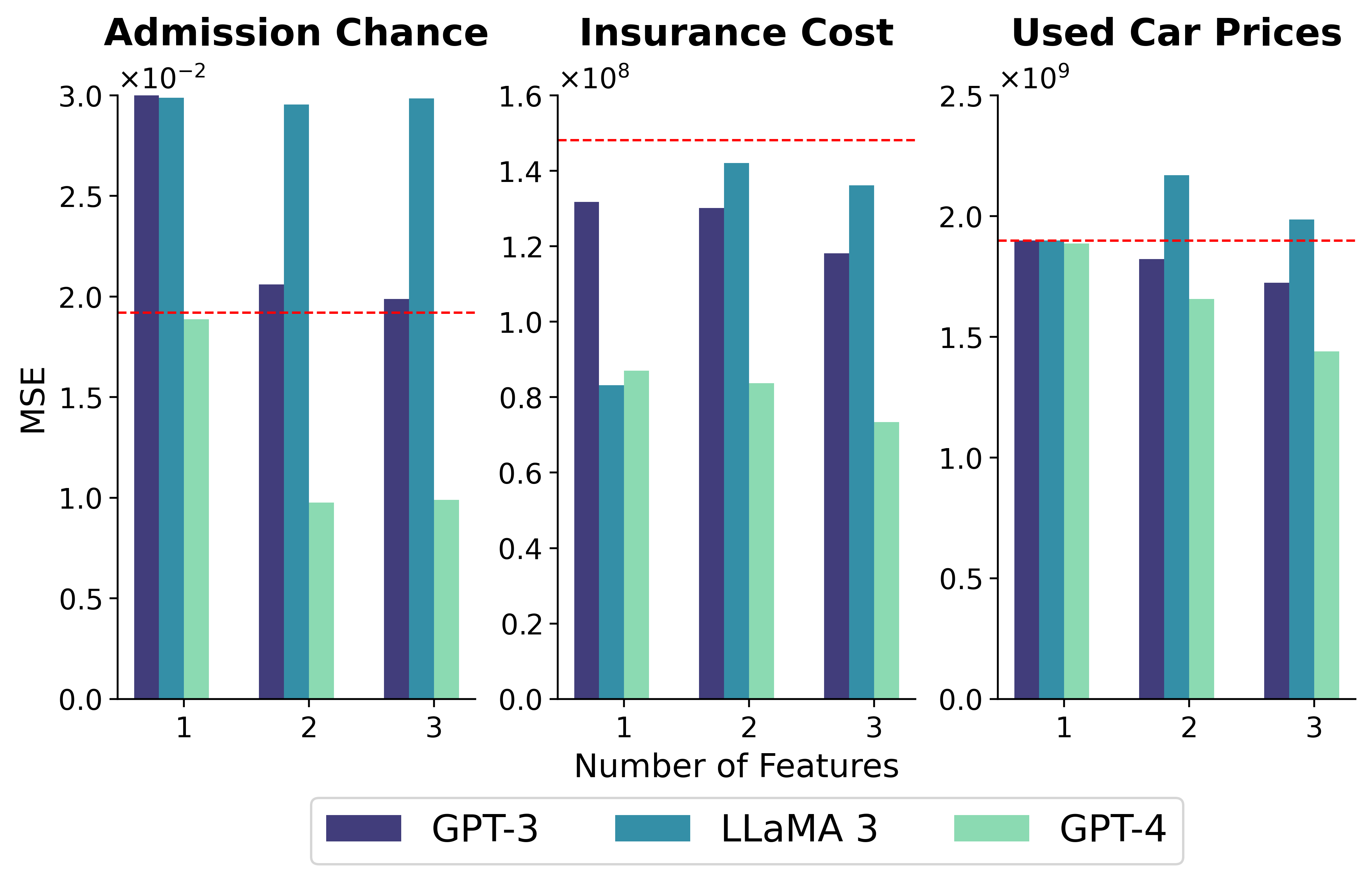}
    \caption{Baseline results (Direct QA configuration) across datasets and number of features. The dashed red line shows the performance of the Mean model. } 
    \label{fig:baselineresults}
\end{figure}

\subsection{Knowledge Retrieval Assessment}

To establish a baseline for our subsequent analyses, we first evaluate the LLMs' performance using only their knowledge retrieval, without any in-context examples. This assessment is required to understand how the in-context examples in other prompt configurations modulate the LLMs' performance and ICL mechanisms. The performance of LLMs with \textit{Direct QA} prompt configuration is shown in Figure~\ref{fig:baselineresults}. The red dashed line shows the performance of Mean model, which outputs the mean of the dataset independent of the input features.As can be observed, the \textbf{additional features generally improve the performance} across tasks (with inconsistencies when using LLaMA 3). However, the LLMs' utilization of features does not directly correlate with feature importance, as explained in Section~\ref{sec:datasets}. As for the impact of datasets, the Admission Chance dataset consistently shows the poorest results, with most outcomes at or above the Mean model's MSE. As mentioned earlier, this dataset is the least exposed to LLMs, which likely explains these findings.

We further explored the Direct QA approach by asking the LLM to explain its reasoning based on the given features before providing the final estimate. This variation yielded predictions comparable to the Direct QA results without offering additional insights and consequently was excluded from our main prompt configurations. See Appendix~\ref{appendix:reasoning} for example answers and related diagrams.


\begin{figure*}[h]
    \centering
    \includegraphics[width=1.0\textwidth]{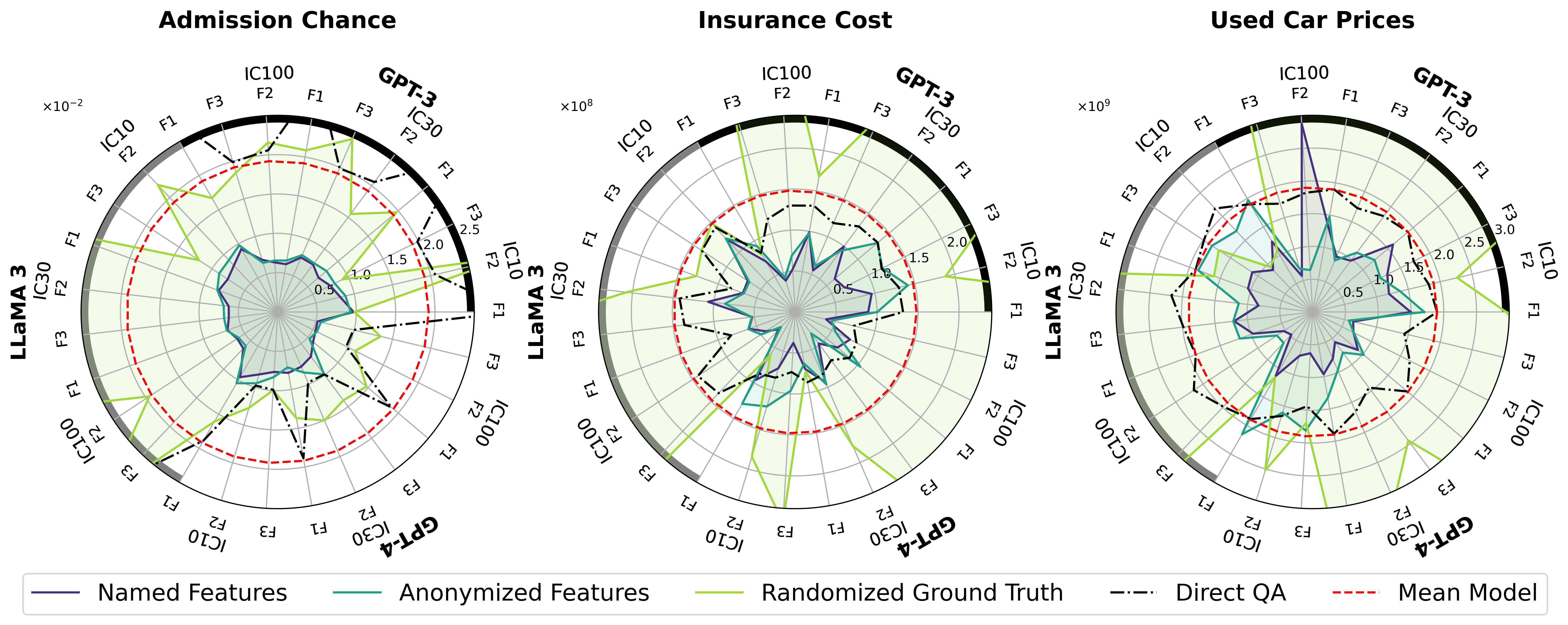}
    \caption{Comprehensive comparison of prompt configurations' effects on our models across various factors, shown in a hierarchy. The top level for each dataset distinguishes between GPT-3, LLaMA 3, and GPT-4 results using black, grey, and white arcs, respectively. The notation $IC_i$ indicates the number of in-context examples, while F1, F2, and F3 represent the use of the first feature, the first two features, and all three features, respectively. The MSE scale of each dataset is shown at top left corner.} 
    \label{fig:configurationdiagrams}
\end{figure*}

\subsection{Learning/Knowledge Retrieval Interplay}

We have defined four main prompt configurations to vary the degree to which the model uses its knowledge retrieval or learning from the in-context examples.
By comparing the performance of these prompt configurations, we can understand how LLMs utilize knowledge retrieval and learning from the outputs. 
Figure~\ref{fig:configurationdiagrams} compares these prompt configurations' effects on LLMs. The \textit{Randomized Ground Truth} prompt configuration, shown with the lime color, consistently yields the worst results. Note that the names of the features are revealed here. As a result, this prompt configuration creates a scenario where patterns in the data may contradict the model's internal knowledge. For instance, in the Insurance dataset, some in-context examples show that people who smoke less, require lower insurance costs, while others indicate higher costs for the same group. The negative impact of this prompt configuration on performance becomes more significant as the number of in-context examples with random outputs increases. This is particularly evident when using 100 in-context examples. These results suggest that the LLMs are using the output variables to learn from the examples, and increasing the number of \textbf{in-context examples shifts the spectrum from knowledge retrieval to learning.} Our findings, which indicate that LLMs can be pushed to prioritize learning from the in-context examples over knowledge retrieval, challenge the uselessness of the outputs claimed by~\cite{min-etal-2022-rethinking,li-etal-2024-language}.

Comparing the \textit{Named Features} and \textit{Anonymized Features} prompt configurations shows the power of combining the two paradigms of learning from in-context examples and knowledge retrieval. Anonymized Features prompt configuration, shown with the green color in Figure~\ref{fig:configurationdiagrams}, allows only usage of the numeric part of the features for learning. This prompt yields better results than the Direct QA and the Mean model. Named Features prompt configuration, which adds the actual name of the features to the examples, is shown in purple. By encouraging the use of knowledge resulting from the added names of the features, it consistently outperforms Anonymized Features prompt configuration across variations of the factors, that is the number of in-context examples and the number of features. These results show that LLMs can exploit \textbf{both in-context examples' outputs for learning and clues like feature names for knowledge retrieval}. Note that when we replaced the feature names with random, unrelated names with the same ranges as the original features (e.g., replacing Smoker Status with Married since both are binary), the results remained the same as those of the Anonymized Features prompt configuration.


\begin{figure*}[h]
    \centering
    \includegraphics[width=1.0\textwidth]{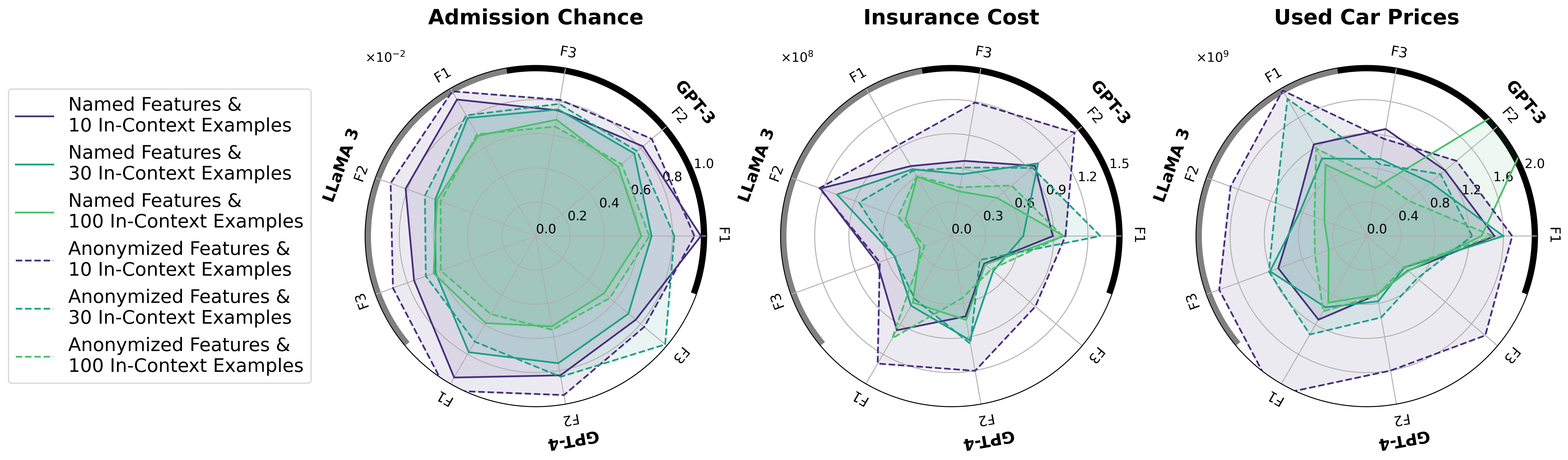}
    \caption{Comparison of the number of in-context examples using Named Features (straight lines) and Anonymized Features (dashed lines) prompt configurations. F1, F2 and F3 indicate using 1st (F1), then 1st and 2nd (F2), and all three features (F3).The MSE scale of each dataset is shown at the top left corner.
} 
    \label{fig:Wholepicture_incontext}
\end{figure*}





\subsection{Knowledge Retrieval Compensates for In-context Examples}

As previously noted, the \textit{Named Feature} prompt configuration, which combines knowledge and learning, generally outperforms the \textit{Anonymized Feature} prompt configuration, which relies only on learning. Regarding these prompt configurations, analyzing our defined factors, such as the number of in-context examples, reveals more insights about ICL. Conventionally, in the realm of LLMs, more in-context examples tend to improve the task outcome. However, as demonstrated in Figure~\ref{fig:configurationdiagrams}, performance deteriorates with more in-context examples when outputs are random, even if the model possesses knowledge about the subject. Figure~\ref{fig:Wholepicture_incontext}, which compares the performance of different numbers of in-context examples, reveals that the Named Feature prompt configuration performs better when fewer in-context examples are used (solid vs dashed lines). A significant performance gap between the two prompt configurations is observed at 10 in-context examples, shown with the solid and dashed purple lines. However, at 100 in-context examples, shown with the solid and dashed lime lines, the performance levels converge. This underscores the potential for \textbf{reducing the number of required in-context examples} by providing task-specific information, thereby shifting the spectrum from learning to knowledge retrieval. 

\begin{figure}[!h]
    \centering
    \includegraphics[width=0.48\textwidth]{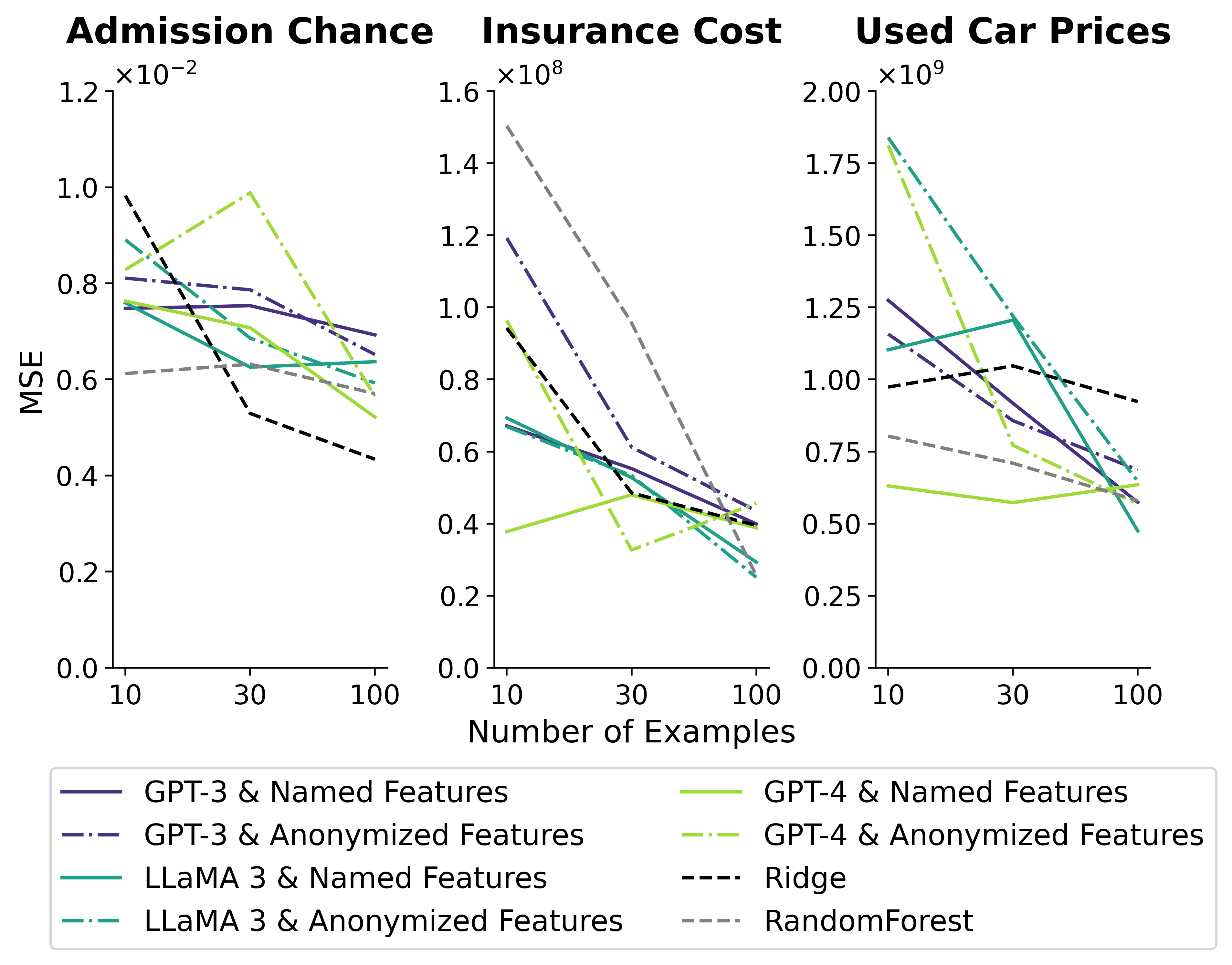}
    \caption{Performance of Named Features and Anonymized Features prompt configurations, Ridge, and RandomForest for 3 features based on the number of in-context examples.} 
    \label{fig:faststart}
\end{figure}

Moreover, Figure~\ref{fig:faststart} shows the advantage of the Named Features prompt configuration over both the Anonymized Features prompt configuration and also the traditional machine learning models by comparing them across various numbers of in-context examples. With fewer in-context examples (30 and especially 10), Named Features prompt configuration models generally outperform all other models. Their results indicate that the type of information in the LLMs' prompt can encourage exploiting their internal knowledge. Moreover, the LLMs' capability to \textbf{retrieve knowledge makes them more data-efficient, i.e., few-shot learners, than even classical machine learning models}, RandomForest and Ridge. However, this advantage applies primarily in a low-data regime and does not account for issues such as the higher training and execution costs of LLMs, which we discuss further in Section~\ref{sec:discussion}. All the models eventually converge at 100 examples when sufficient data for regression is provided.


\subsection{More Features Help Knowledge Retrieval}

Unlike the number of in-context examples, the number of features stands out as a unique factor that can improve \textbf{both learning and knowledge retrieval} in LLMs. As additional features are incorporated into the context, LLMs can better learn the relationships between features and outputs and also retrieve more relevant knowledge. This clearly helps in the case of Named Features prompt configuration.

\begin{figure}[!h]
    \centering
    \includegraphics[width=0.48\textwidth]{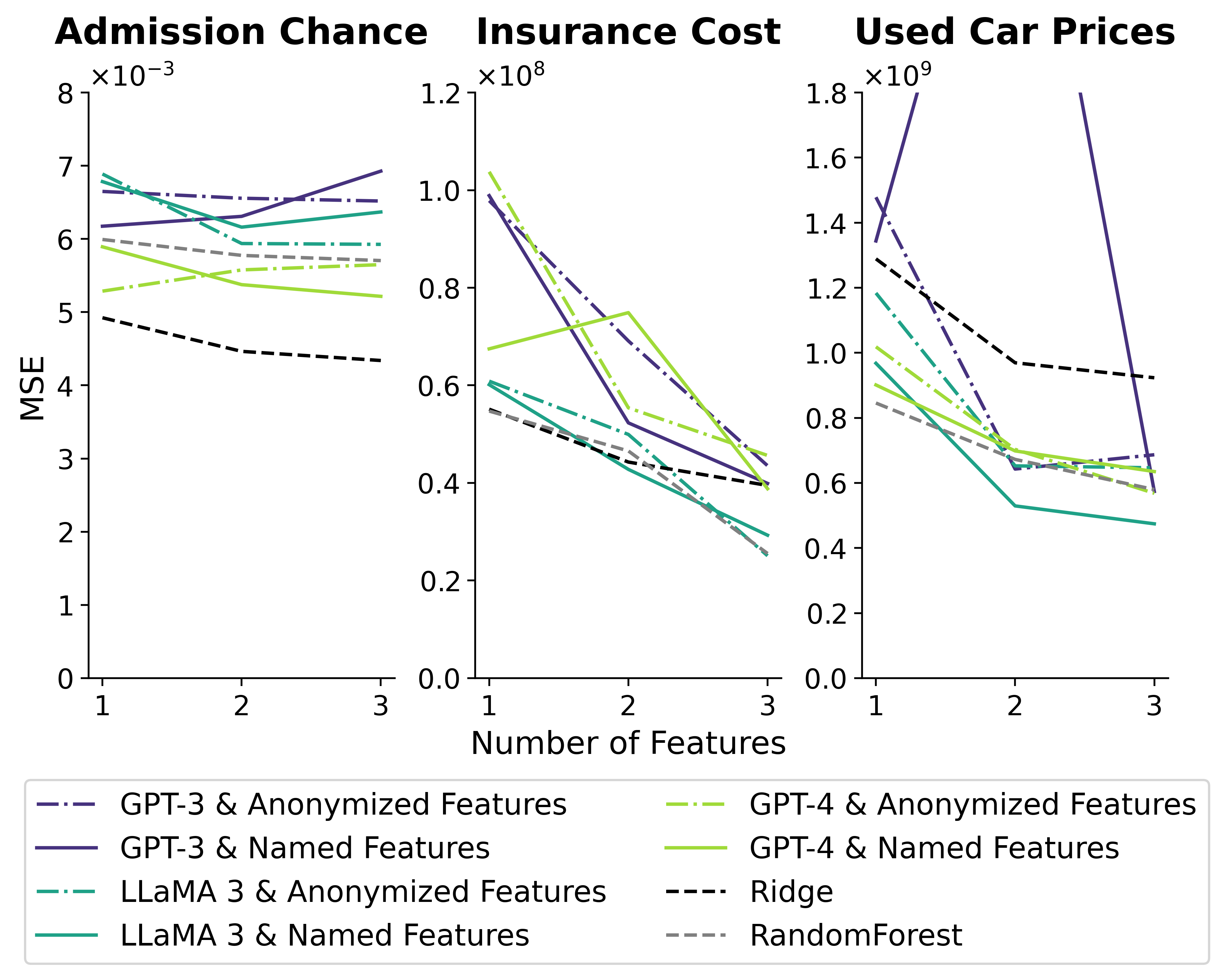}
    \caption{Performance of Anonymized Features prompt configuration, Ridge, and RandomForest for 100 in-context examples based on the number of features.} 
    \label{fig:no_name_100_vs_ridge}
\end{figure}

However, the less obvious insights in this section emerge from the \textit{Anonymized Features} prompt configuration, which relies only on learning and consistently improves with the addition of features. As shown in Figure~\ref{fig:no_name_100_vs_ridge} for 100 in-context examples (the dotted lines), each additional feature either improves the results or maintains near-optimal performance. A similar, but less robust, trend is seen in the 30 in-context examples' diagram in Appendix~\ref{appendix:reasoning}. While adding more features improves results, this improvement is more significant in language models compared to machine learning models such as Ridge and RandomForest in Figure~\ref{fig:no_name_100_vs_ridge}. Given that in this prompt configuration 1) feature names are anonymous, 2) only raw numerical data is available for output estimation, and 3) the improvement in the \textbf{performance of LLMs does not correlate with the importance} of the features, we conjecture that the results could be influenced by \textbf{data contamination from the observed numbers.} This assertion is supported by the fact that this phenomenon is absent in the Admission Chance dataset, the least likely seen dataset by LLMs.

In the Named Feature prompt configuration for 100 in-context examples depicted in Figure~\ref{fig:no_name_100_vs_ridge}, the results show an overall improvement, yet the trend is not as consistently downward or smooth as observed in the Anonymized Features prompt configuration when features are added. This fluctuation suggests that the knowledge retrieval aspect becomes more prominent and occasionally supersedes the learning component as the number of features increases. Based on these observations that happen at a high number of 100 in-context examples, we hypothesize that adding features \textbf{primarily enhances the knowledge retrieval aspect rather than contributing to the learning}.

\subsection{Quantitative Analysis}

\begin{figure}
    \centering
    \includegraphics[width=1.0\linewidth]{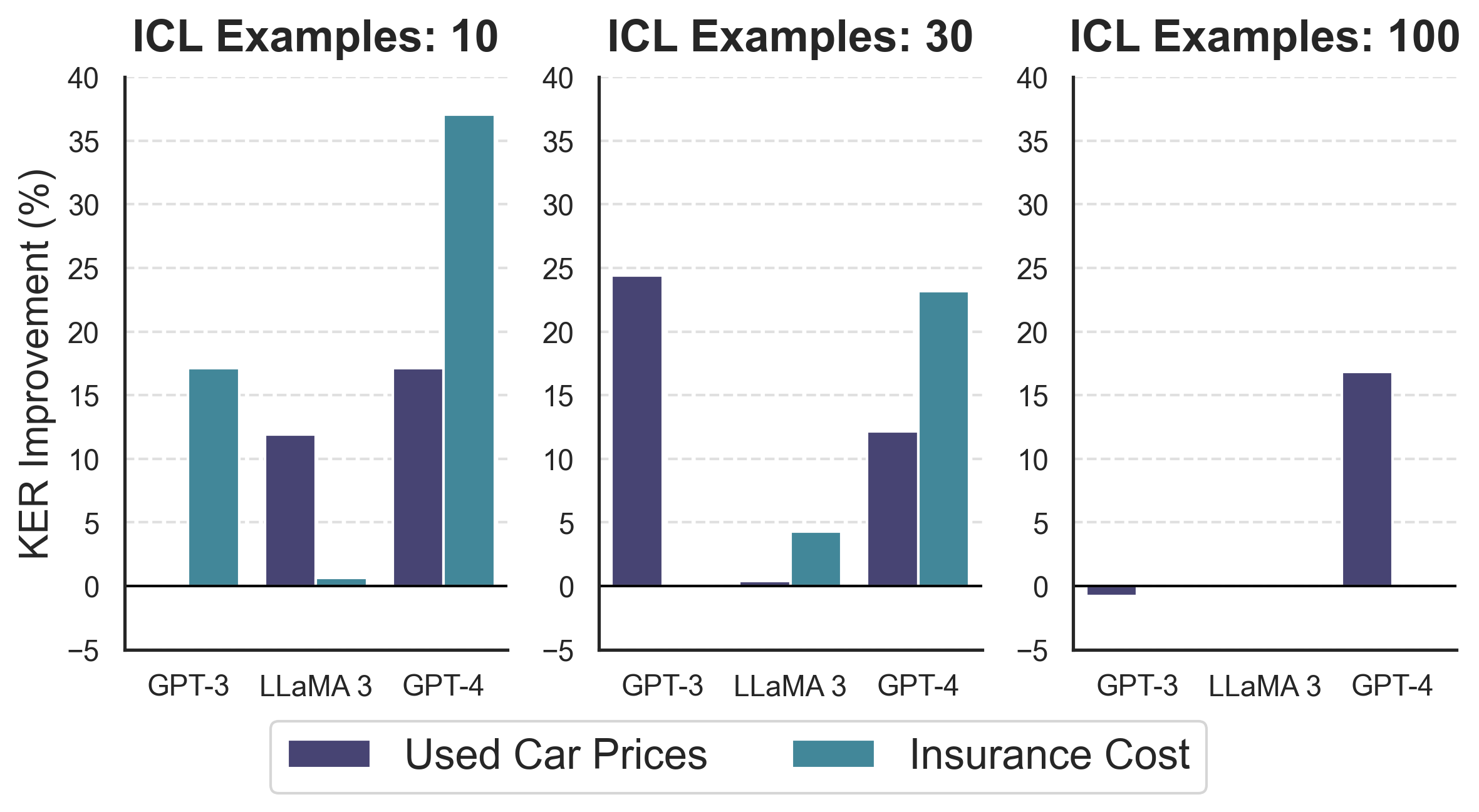}
    \caption{KRER improvement across different models and datasets with varying numbers of in-context examples. The improvements become more notable as the number of in-context examples decreases.}
    \label{fig:QAnalysis}
\end{figure}

Following \cite{yoo-etal-2022-ground}, we introduce a quantitative metric to assess the impact of knowledge on model performance. We denote the rate of improvement of a prediction after the addition of knowledge (named features) as \textit{Knowledge Effect Ratio} (KER), which can be calculated as such:

\begin{equation}
\texttt{KER} = \frac{|Y_{AF} - Y_{GT}| - |Y_{NF} - Y_{GT}|}{|Y_{AF} - Y_{GT}|} \times 100
\end{equation}

where $Y_{AF}$ shows a prediction from the Anonymized Features Configuration, $Y_{NF}$ shows a prediction from the Named Features Configuration and $Y_{GT}$ is the ground truth value. This formula measures the rate of change in error (Mean Absolute Error) due to added knowledge in learning. We compute the KER for each dataset and combination of factors and then select its median improvement rate to mitigate the influence of outliers. As expected, this rate remains near zero across almost all factor combinations for the Admission Chance dataset, given that it does not benefit from knowledge in our tests. For the other two datasets, Figure~\ref{fig:QAnalysis} shows the average improvements across feature factors for each dataset and the number of in-context examples. Notably, the most significant gains emerge when the number of in-context examples is small, aligning with our earlier observations.

\section{Discussion}
\label{sec:discussion}

Our evaluation framework and extensive experiments provide insights into the ICL mechanisms of LLMs. In this section, we discuss the implications of our findings, address limitations, and suggest directions for future research.

\subsection{Controlling ICL Mechanisms}

Understanding and manipulating ICL mechanisms in LLMs proves invaluable for practical applications and effective prompt engineering. Our experiments show that the LLMs use both learning from the in-context input-output examples and prior information with varying extents. We found that increasing the number of in-context examples encourages learning while adding more features \textbf{mainly boosts knowledge retrieval}. However, these benefits only materialize when there is room for improvement in the respective mechanism. For example, using more than 100 in-context examples did not help LLMs, and in the Admission Chance dataset, where the features were highly correlated, the addition of features did not help the performance either. This insight explains why \cite{min-etal-2022-rethinking}'s tasks appeared not to utilize output labels for learning; In their tasks, the dominant knowledge retrieval aspect likely eclipsed any learning benefits. Our findings also challenge the view of LLMs as merely meta-learners. While previous research demonstrates decoder models' capacity for meta-learning, the complex interplay of different training objectives (unknown in some proprietary models) and other factors introduce a significant knowledge retrieval component. Consequently, earlier findings can only be directly extrapolated to LLMs when considering all these dynamics.

\subsection{Practical Applications}

Our findings emphasize the importance of striking an optimal balance between the number of meaningful features and the number of in-context examples to optimize LLMs performance. By strategically reducing the quantity of in-context examples while increasing the number of named features, it is possible to achieve resource efficiency 
and other potential benefits, such as mitigating data biases, without compromising performance. Conversely, when tackling tasks unfamiliar to the model, pruning less important features can free up space in the LLM’s token context, allowing it to accommodate more in-context examples. This shifts the ICL mechanism towards the learning side of the spectrum and improves the model's performance. 

\subsection{Traditional ML Models vs LLMs}

Our findings show that LLMs can be more data-efficient in low-data regimes, leveraging their pre-trained knowledge to achieve strong performance with only a few in-context examples. This advantage holds against traditional ML models even with the addition of more than 3 features, given that in classical ML models, with sparse data, more features lead to overfitting- a problem we encountered on our datasets as well. In addition, in our experiments, LLMs excelled at identifying the most important features of the datasets, performing on par with the classical feature selection method we used. On the negative side, LLMs require substantial computational resources to train and deploy. Meanwhile, traditional models such as Ridge regression or RandomForest can be trained rapidly on standard hardware and may eventually surpass LLMs if enough data is provided.

\subsection{Data Contamination}

In our experiments, we showed that LLMs can perform regression on realistic datasets, even when feature names are anonymized. However, as detailed in the experiments section and illustrated in Figure~\ref{fig:no_name_100_vs_ridge}, the improvement does not correlate with the actual importance of input features in the dataset. This discrepancy strongly suggests that data contamination occurs even at the level of numerical values. 
These findings challenge the claims made by \cite{vacareanu2024words}, which uses a prompt configuration similar to Anonymized Features to avoid data contamination. Their study utilizes either well-known Friedman formulas or formulas that can be closely approximated (using two decimal places) by common mathematical expressions\footnote{E.g. $y = 10x + \sin(5\pi x) + \cos(6\pi x) \approx 10x$.}.

\subsection{Order of Features \& In-Context Examples}

In our experiments, we explored two factors: rearranging the order of features in the Direct QA prompt configuration and sorting the in-context examples in the Anonymized Features prompt configuration. When permuting the order of our three features (resulting in six possible permutations), we found only a slight change in performance. After sorting the in-context examples by their label values (with three features), we tested the impact on performance using sets of 10 and 100 in-context examples. While both scenarios resulted in a drop in performance, the drop was notably steeper when using 100 examples. Among the models we evaluated, GPT-4 was the least affected, maintaining robust performance with 10 sorted examples but showing a marked decrease with 100 examples.

We hypothesize that sorting examples by label values may cause LLMs to prioritize the sorted label pattern rather than the relationships between features and labels—an effect possibly affected by LLMs’ inherent ability to detect and continue sorting patterns~\cite{Graph_of_thoughts}. We noticed that when the examples were sorted in ascending order, the mean predicted value went up, and when sorted in descending order, it went down. This suggests that the sorting pattern guides the model’s predictions, and it is best to keep the order of in-context examples randomized.

\section{Conclusion}

In this study, we have explored the ICL capabilities of LLMs through an empirical study on regression tasks. Our results demonstrate that LLMs utilize a blend of retrieved internal knowledge and learned information from in-context examples. Our findings extend the evaluations of prior hypotheses on ICL. For example, we evaluate the usage of outputs in in-context learning examples and show how to manipulate their effectiveness. This work not only advances our understanding of LLMs' in-context learning phenomenon but also offers practical insights for optimizing their application through careful prompt engineering.

\section*{Acknowledgments}

This project is partially supported by the Office of Naval Research (ONR) grant N00014-23-1-2417. Any opinions, findings, and conclusions or recommendations expressed in this material are those of the authors and do not necessarily reflect the views of Office of Naval Research. We thank the anonymous reviewers for their valuable feedback and suggestions, which have helped improve this work.

\section*{Limitations}

\subsection*{Scope of Study}

Our framework focuses on regression tasks in line with most previous meta-learning research. We also utilize three diverse regression datasets. More regression (or classification) datasets with different characteristics, such as different relationships between their features, can be used for future investigation, which could elaborate more on the findings of our work with the previous work that dismissed the importance of outputs for learning.

\subsection*{Interpretability}

We interact with LLMs as a black box and the underlying neural mechanisms remain opaque in our experiments. Integrating this approach with interpretability techniques could yield a deeper understanding of ICL.

\subsection*{Experimental Constraints}

The token limit of some LLMs prevented us from testing beyond 100 in-context examples and three features in certain combinations. Although we found that results with 200 in-context examples closely mirrored those with 100, including a 4th feature could have further reinforced our findings. When tested, the results of the 4th feature aligned with our conclusions. However, we ultimately restricted our analysis to three features to maintain a comprehensive combination across all models. 


\subsection*{Data Contamination Challenge}

Addressing the challenge of data contamination remains a complex issue~\cite{sainz-etal-2023-nlp,balloccu-etal-2024-leak}. As mentioned earlier, it is hard to distinguish the degree of Data Contamination from approximate Knowledge Retrieval. Based on our findings, data contamination can happen even with the sequence of numerical values without any linguistic clues. To mitigate this issue, we suggest using distributions that are neither widely recognized nor easily approximated by common mathematical expressions. Furthermore, for datasets, it is advisable to utilize information sourced from regions beyond the United States or outside the primary training domain of the language model. This approach helps to minimize the potential influence of knowledge retrieval in generating the results.

\bibliography{anthology_0,anthology_1,custom}

\newpage
\appendix

\section{Related Work}
\label{appendix:related_work}

As mentioned in the introduction, the research in this field can be categorized into meta-learning~\cite{schmidhuber1987evolutionary} and knowledge retrieval. So far, most works are theoretical, and most can be categorized into one of these two groups. However, both of these aspects are changing as this field rapidly expands; New empirical research is coming along, and new ideas are emerging, such as ours, that can not easily fit into one of these groups. For example, \cite{li2023dissectingchainofthoughtcompositionalityincontext} proposes a structure that improves a basic transformer's in-context learning capabilities by breaking the input into multiple steps. The transformer still uses the outputs but also the knowledge that is given to it \textbf{in its input}. Another example would be ~\cite{kossen2024incontextlearninglearnslabel}, which empirically tests LLMs and claims that they can learn from the output labels but in an arbitrary manner, a challenge our paper aims to address. Another interesting example that we cannot simply put into a group is ~\cite{NEURIPS2023_cda04d7e}, which performs meta-in-context learning by showing similar entire tasks and their solutions in the context of the model. Despite these emerging works, it's worth noting that most papers can still be classified into our established groups. When evaluating the literature, it's crucial to distinguish between theoretical claims and those based on empirical or synthetic data. Some theoretical papers suggest or imply that their findings might extend to LLMs, but such extrapolations should be approached with caution, as highlighted by~\cite{deutch-etal-2024-context,shen2024pretrainedtransformerslearnincontext}.

In the meta-learning research, \cite{bai2023transformers} theoretically prove that transformers can be trained to implement linear models within a specific error threshold. Then, they practically test the capability of an encoder-based transformer to implement generalized linear models in context and show that they are also robust against noisy data. In order to test the ICL capability of transformers, \cite{Garg2022WhatCT} use a GPT-2 architecture which is closer to an LLM than a simple transformer. However, they modify this GPT-2 architecture for regression using the input features inside the embeddings instead of using tokens and output a number instead of generating tokens. Their model's performance surpasses linear models as well as KNN models, XGBoost, and a 2-layer neural network. Different from these, ~\cite{vonoswald2023uncoveringmesaoptimizationalgorithmstransformers} hypothesize that the strong performance of Transformers comes from an architectural bias towards mesa-optimization which they test on simple sequencing tasks, and \cite{cheng2024transformersimplementfunctionalgradient} suggest Transformers can implement gradient descent in function space, enabling them to learn linear and non-linear models.

In the realm of empirical research, our focus is exclusively on studies examining ICL, distinct from related works involving numerical concepts such as numerical reasoning~\cite{razeghi-etal-2022-impact}. \cite{vacareanu2024words} uses a lot of different LLMs such as Gemini-pro~\cite{geminiteam2024gemini15unlockingmultimodal}, Claude 3~\cite{claude3}, GPT-4~\cite{openai2023gpt4} to test their ICL regression capabilities. They use various settings with up to three features, but not all features are relevant to the output as limited as a result. This work conducts rigorous testing with many models, and the only issue it has is regarding its claim of avoiding data contamination, which is not well-supported. They either use well-known Friedman formulas or formulas that can be closely approximated (using two decimal places) by common mathematical expressions such as $y = 10x + \sin(5\pi x) + \cos(6\pi x)$ simplified as $y = 10x$. Also, the data is based on formulas, which are not realistic regression datasets.

The first paper that argues output labels do not matter was ~\cite{min-etal-2022-rethinking}. It tested various LLMs across multiple tasks and concluded that replacing the outputs with random labels does not affect the results, and the only important elements of ICL are limited to defining the label space, input distribution, and overall task format.
Following this work, \cite{pan-etal-2023-context} uses the same sentiment analysis, toxicity detection, natural language inference/paraphrase detection, and topic/stance classification tasks but argues that learning and retrieval occur under different circumstances. In contrast to our research, their approach employs less complex tasks (same tasks used in ~\cite{min-etal-2022-rethinking}), which they acknowledge could influence the observed patterns. Additionally, they utilize more basic models, including earlier and smaller iterations of GPT-3 such as ada. As a result, their findings yield distinct trends and visual representations compared to our work. Inspired by ~\cite{hahn2023theoryemergentincontextlearning}, 
  \cite{li-etal-2024-language} dismisses the significance of output labels and proposes that learned solutions can be combined for inference. The main limitation of this work is the exclusive use of LLaMA 2~\cite{touvron2023LLaMA}. Our experimental results differ from~\cite{min-etal-2022-rethinking,pan-etal-2023-context,li-etal-2024-language}, which requires a unified platform for direct comparison, which is practically infeasible. We argue this conflict is mainly due to the task and model selection. We will explain further on \cite{li-etal-2024-language} in Appendix~\ref{appendix:aclpaper}.

\section{Learning from Outputs}
\label{appendix:aclpaper}

In this section, we examine the paper \cite{li-etal-2024-language} and why its findings differ from ours. As mentioned earlier, the main problem with this work lies in its exclusive use of LLaMA 2~\cite{touvron2023LLaMA}, a model known to struggle with long token contexts~\cite{machlab2024llmincontextrecallprompt,meta_LLaMA_issue_887} which negatively impacts the ICL. While the choice of LLaMA 2 as a powerful open-source LLM at the time is understandable, LLaMA 3~\cite{meta2024LLaMA3}, which addressed many of LLaMA 2's issues and offered significantly improved capabilities, was already available at the time of publication. Some of their experiments require direct access to the LLM architecture, which necessitates an open-source LLM. However, many of their experiences can be done with the GPT family or other powerful models. Based on the LLM choice alone, this work is severely limited. Further limiting the study, the authors selected three simple review sentiment/news-type datasets that even basic machine learning models can excel at.

Regarding specific experiments, experiment 4.1 demonstrates that mapping each label to another fixed label yields results worse than a random baseline, even with 40 in-context examples. This finding contradicts our work and other related studies, which have shown that models can learn from outputs. We suspect this discrepancy may be attributed to LLaMA 2's well-documented issue of forgetting tasks as the number of tokens increases. Confusion ensues because the outputs likely appear nonsensical to LLaMA 2, and it has forgotten its instructions. It would have been beneficial for the authors to specify whether LLaMA 2 was outputting incorrect labels or entirely irrelevant answers.

Experiments 4.2 and 5.1 also raise questions, as they rely on token embeddings, which, without further evidence, appear unrelated to ICL. For example, \cite{li2023dissectingchainofthoughtcompositionalityincontext} suggests that the level-by-level inference in transformers is necessary for ICL. The combination of embeddings and other factors, such as training the linear model for 80 epochs to match LLaMA 2's 80 layers, seems arbitrary and lacks clear justification in these experiments.

In conclusion, while the paper \cite{li-etal-2024-language} offers interesting insights specifically regarding its categorization of the three hypotheses, its methodological choices and reliance on LLaMA 2 significantly limit its relevance to the broader discussions of in-context learning in language models.

\section{Datasets}
\label{appendix:datasets}

To assess the LLMs' performance across diverse domains and complexity levels, we select three regression datasets. These datasets are pre-processed to select their most important features. All numerical values in the datasets are rounded to two decimal points. We divide each dataset into two splits: an in-context subset comprising 100 instances and a test subset with 300 instances.

\begin{figure}[h]
    \centering
    \includegraphics[width=0.48\textwidth]{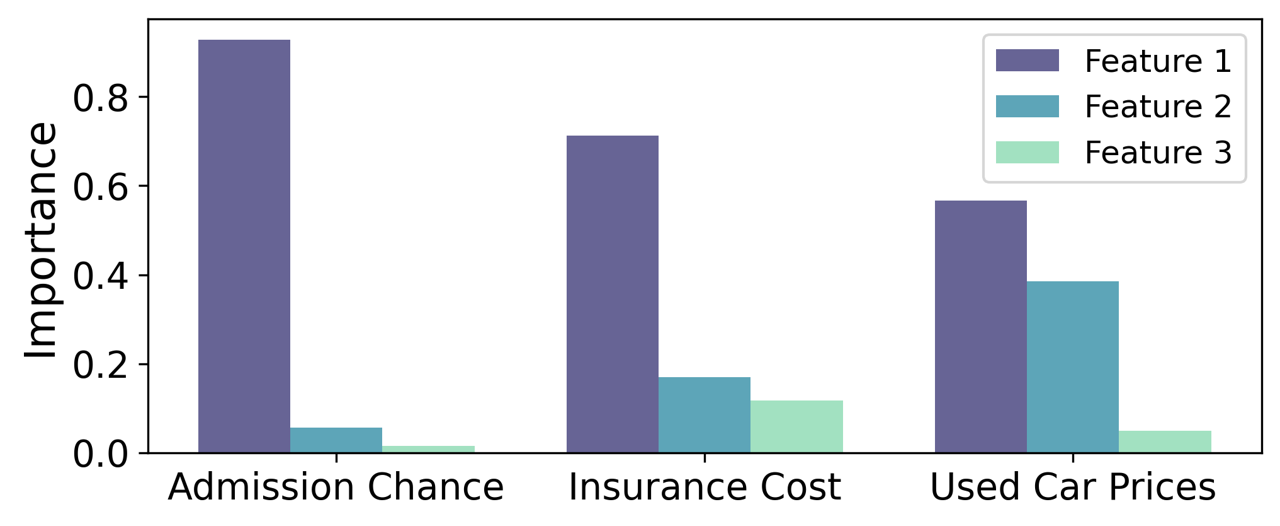}
    \caption{The relative importance of features across our datasets calculated with RandomForest (also reflecting improvements in our Ridge as later shown in Figure~\ref{fig:no_name_100_vs_ridge}.} 
    \label{fig:feature_importance}
\end{figure}

\subsection{Admission Chance} This dataset estimates the \textit{likelihood of admission to graduate programs for Indian students}~\cite{chanceofadmission8862140}. Since this dataset is about Indian students, it may be less seen in our model's training data, reflecting the general imbalance that skews towards USA-centric data~\cite{zhu2023extrapolatinglargelanguagemodels}. The dataset's selected features are CGPA (Cumulative Grade Point Average), GRE Score, and TOEFL Score. Among our three datasets, this is the only dataset with high inter-correlation among each feature pair (Pearson correlation~\cite{pearson1895vii} greater than $0.80$). As a result, the feature importances, which are calculated with RandomForest~\cite{louppe2013understanding} and shown in Figure~\ref{fig:feature_importance}, assign minimal importance to the second and third features. The target variable has a mean of $0.72$ and a standard deviation of $0.14$.

\subsection{Insurance Cost} This dataset focuses on predicting a similar distribution of the \textit{annual individual medical costs billed by health insurance companies in the USA}, drawing from demographic statistics compiled by the U.S. Census Bureau~\cite{lantz2013machine}. The selected features are Smoker Status, BMI, and Age with their importances shown in Figure~\ref{fig:feature_importance}. Smoking Status emerges as the most critical feature, followed by BMI, and then Age. The average and standard deviations of costs are $13,270.42$ and $12,110.01$, respectively.

\subsection{Used Car Prices} We use selected subset of the used car prices dataset~\cite{mital2023ususedcars} that involves predicting the \textit{price of a used Toyota or Maserati car in 2019}. The features include City Fuel Economy, Mileage with similar high importance, and Passenger Car Classification with minimal importance, as shown in Figure~\ref{fig:feature_importance}. The target variable has a mean of $42,279.49$ and a standard deviation of $50,014.51$.

\section{Hyper-Parameters}
\label{appendix:Hyper_Parameters}

As mentioned earlier, we evaluate several LLMs, which are LLaMA 3 70B~\cite{meta2024LLaMA3}, GPT3.5~\cite{brown2020languagemodelsfewshotlearners}, and GPT4~\cite{openai2023gpt4}. We also used Ridge regression~\cite{hoerl1970ridge} and RandomForest~\cite{Breiman2001}. The GPT-3 and GPT-4 versions we use are ''gpt-4-0125-preview``, ''gpt-3.5-turbo-0125``. In these models, the temperature is set to $0.1$, and make tokens is set to $10$. Each time the models do not generate a number (i.e., generate a disclaimer that says ``the data is insufficient''), the seed value, which is initialized to $100$, is increased by one. This precaution was not necessary for GPT-4, but GPT-3.5 sometimes refused to answer the questions. For LLaMA 3, top p value is set to $0.99$, max tokens to $6$, and the temperature to $0.1$. Similar to GPT models, we propmt LLaMA 3 again if it fails to generate a number. For Ridge, the default Python class of scikit\cite{scikit-learn} has the alpha value ( the normalizing variable) of $1$. For RandomForest, however, we wanted to make it more generalized. Consequently, we used $10,000$ estimators and a max depth of 2 for it.

\section{MAE and $R^2$}
\label{appendix:other_metrics}

In this section, we depict the main diagrams in the paper with MAE and $1-R^2$ metrics instead of MSE. MAE is less sensitive to outliers compared to MSE and is expressed in the same units as the original data. $R^2$, also known as the coefficient of determination, ranges from 0 to 1 and represents the proportion of variance in the dependent variable that is predictable from the independent variable(s). An $R^2$ value of 1 indicates that the model explains all the variability of the response data around its mean, while 0 means the model explains none of the variability. It can also be negative, indicating that the model fits the data worse than the Mean model. $R^2$ differs from MSE and MAE because the higher, the better. As a result, to make $R^2$ diagrams comparable, instead of $R^2$, we show the results for $1-R^2$ (1 is included to show the negative results. This also moves the baseline results of $R^2$ to $1$ for the Mean model.

\begin{figure}[h]
    \centering
    \includegraphics[width=0.45\textwidth]{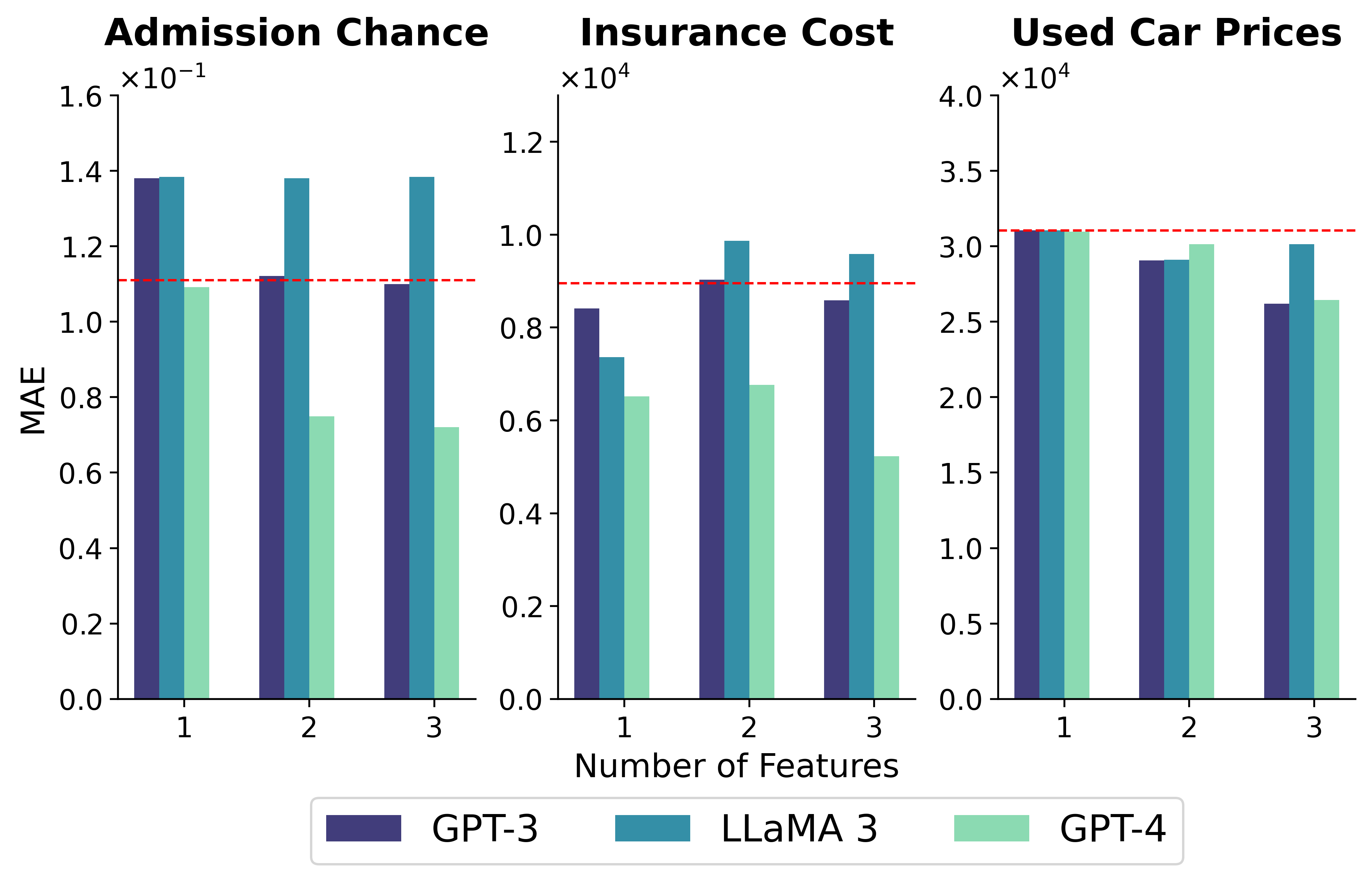}
    \caption{This figure shows Figure 2 of the main paper with MAE instead of MSE. The figure shows the baseline results of LLMs' (Direct QA configuration) across datasets and number of features. The dashed red line shows the performance of the Mean model. } 
    \label{fig:baseline_mae}
\end{figure}

\begin{figure}[h]
    \centering
    \includegraphics[width=0.45\textwidth]{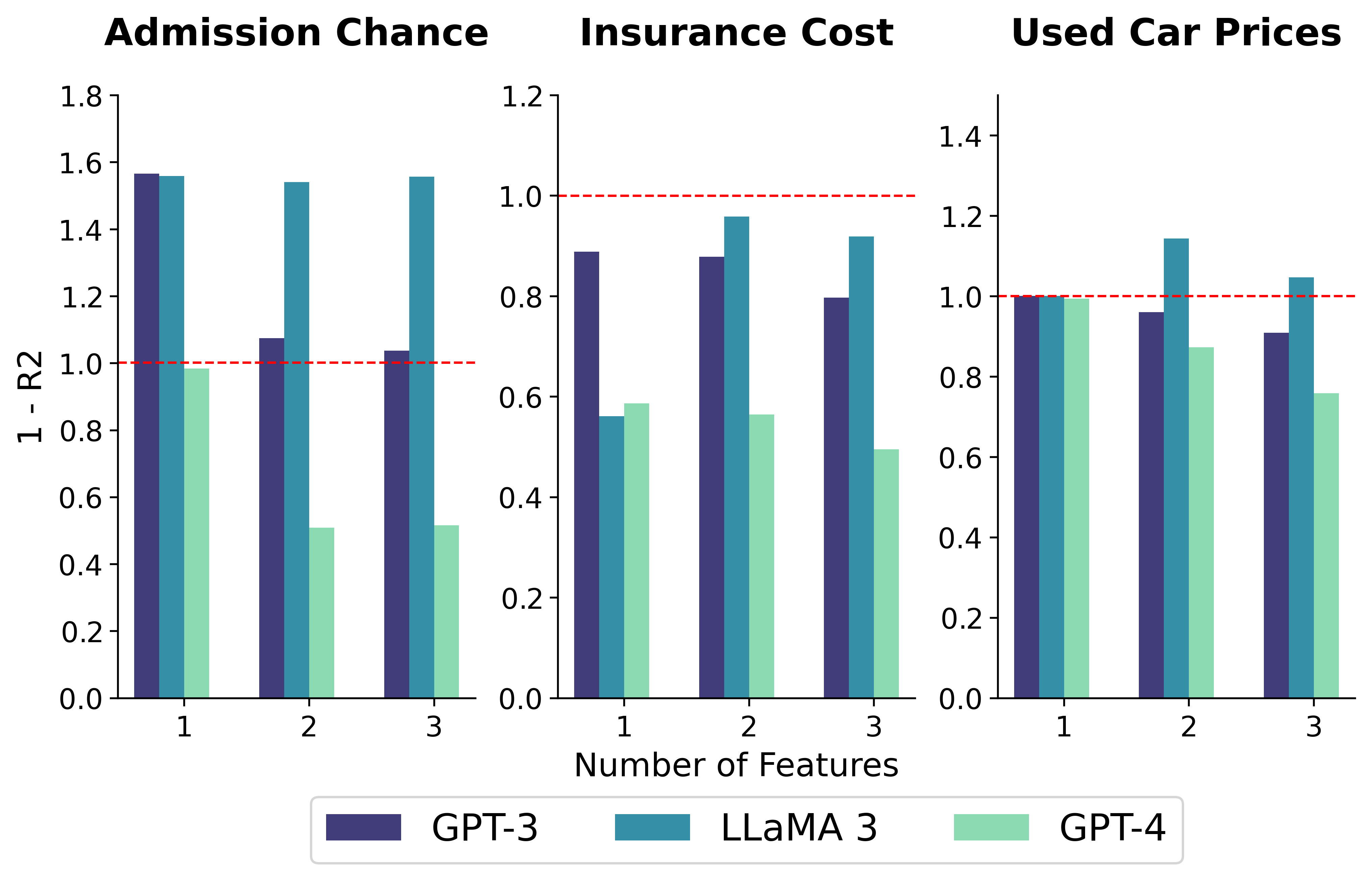}
    \caption{This figure shows Figure 2 of the main paper with 1-$R^2$ instead of MSE. The figure shows the baseline results of LLMs' (Direct QA configuration) across datasets and number of features. The dashed red line shows the performance of the Mean model. } 
    \label{fig:baseline_r2}
\end{figure}

\begin{figure*}[!h]
    \centering
    \includegraphics[width=1.0\textwidth]{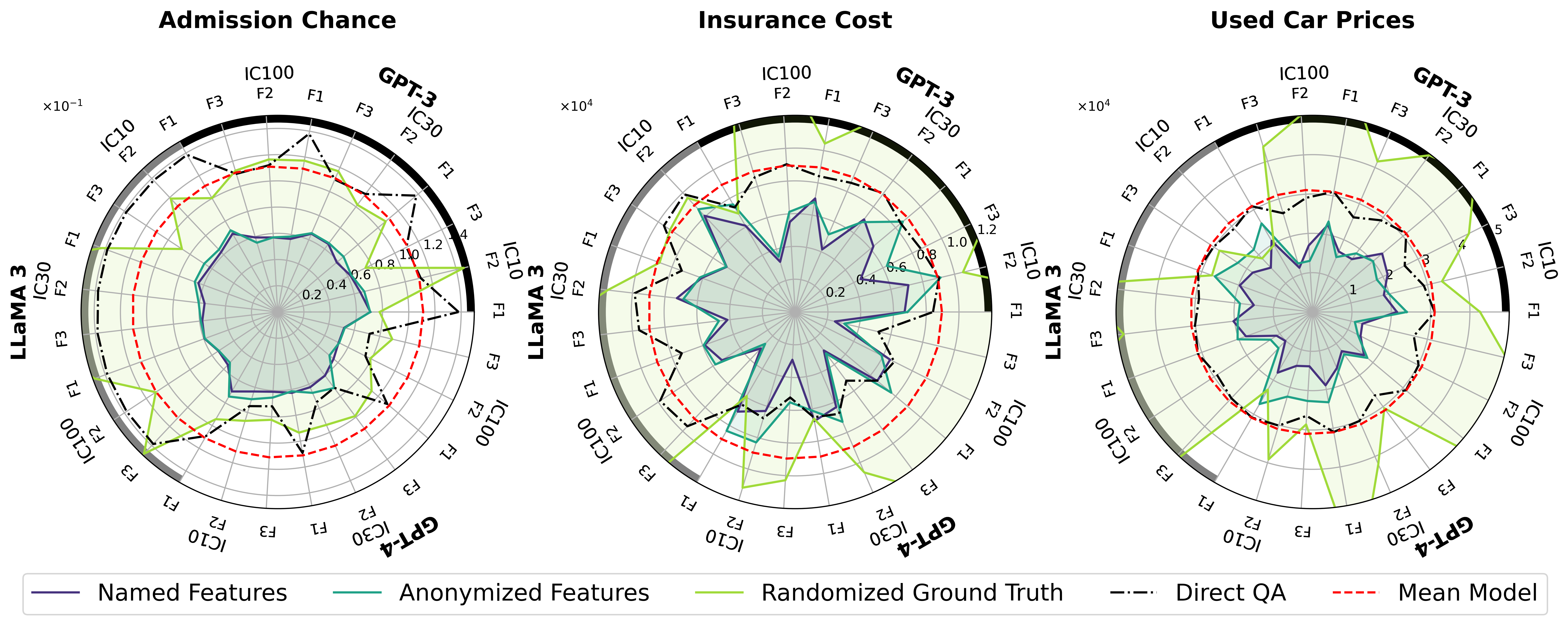}
    \caption{This figure shows Figure 3 of the main paper with MAE instead of MSE. The figure shows the comprehensive comparison of prompt configurations' effects on our models across various factors, shown in a hierarchy. The top level for each dataset distinguishes between GPT-3, LLaMA 3, and GPT-4 results using black, grey, and white arcs, respectively. The notation $IC_i$ indicates the number of in-context examples, while F1, F2, and F3 represent the use of the first feature, the first two features, and all three features, respectively.} 
    \label{fig:wholepicture_mae}
\end{figure*}

\begin{figure*}[!h]
    \centering
    \includegraphics[width=1.0\textwidth]{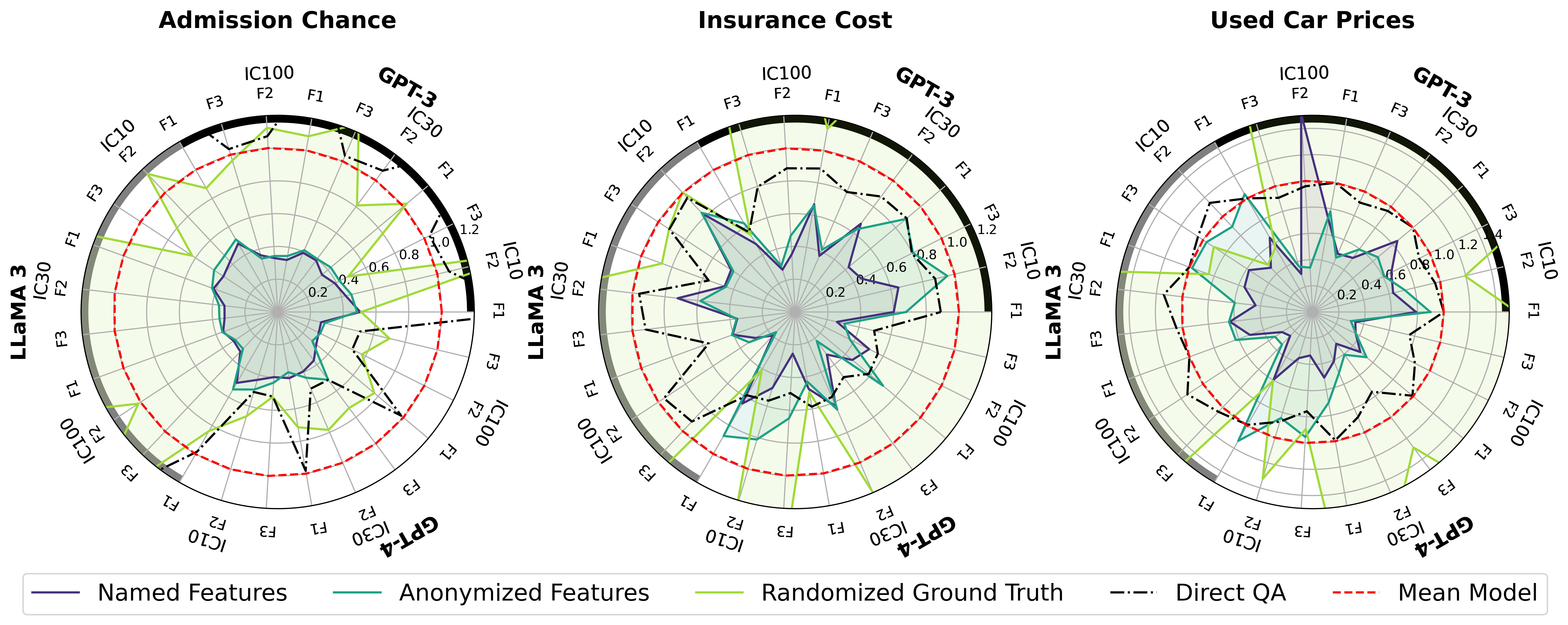}
    \caption{This figure shows Figure 3 of the main paper with 1-$R^2$ instead of MSE. The figure shows the comprehensive comparison of prompt configurations' effects on our models across various factors, shown in a hierarchy. The top level for each dataset distinguishes between GPT-3, LLaMA 3, and GPT-4 results using black, grey, and white arcs, respectively. The notation $IC_i$ indicates the number of in-context examples, while F1, F2, and F3 represent the use of the first feature, the first two features, and all three features, respectively.} 
    \label{fig:wholepicture_r2}
\end{figure*}

\begin{figure*}[h]
    \centering
    \includegraphics[width=1.0\textwidth]{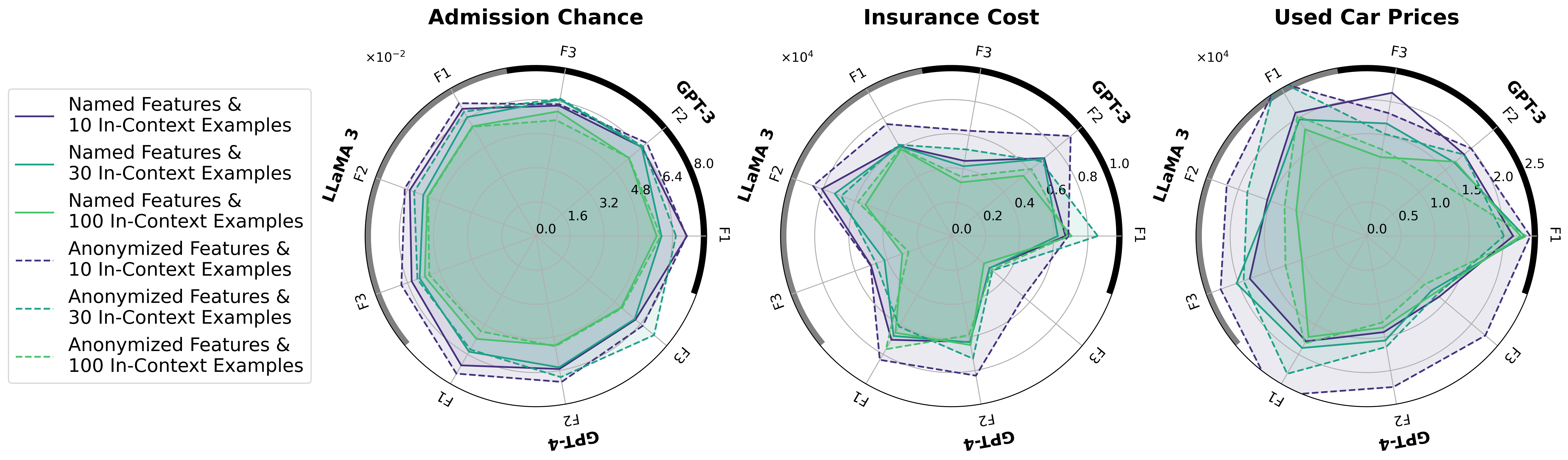}
    \caption{Figure 4 of the main paper with MAE instead of MSE. The figure shows the comparison of the number of in-context examples using Named Features (straight lines) and Anonymized Features (dashed lines) prompt configurations. F1, F2 and F3 represents using 1st (F1), then 1st and 2nd (F2), and all three feature (F3).
} 
    \label{fig:whole_context_mae}
\end{figure*}

\begin{figure*}[h]
    \centering
    \includegraphics[width=1.0\textwidth]{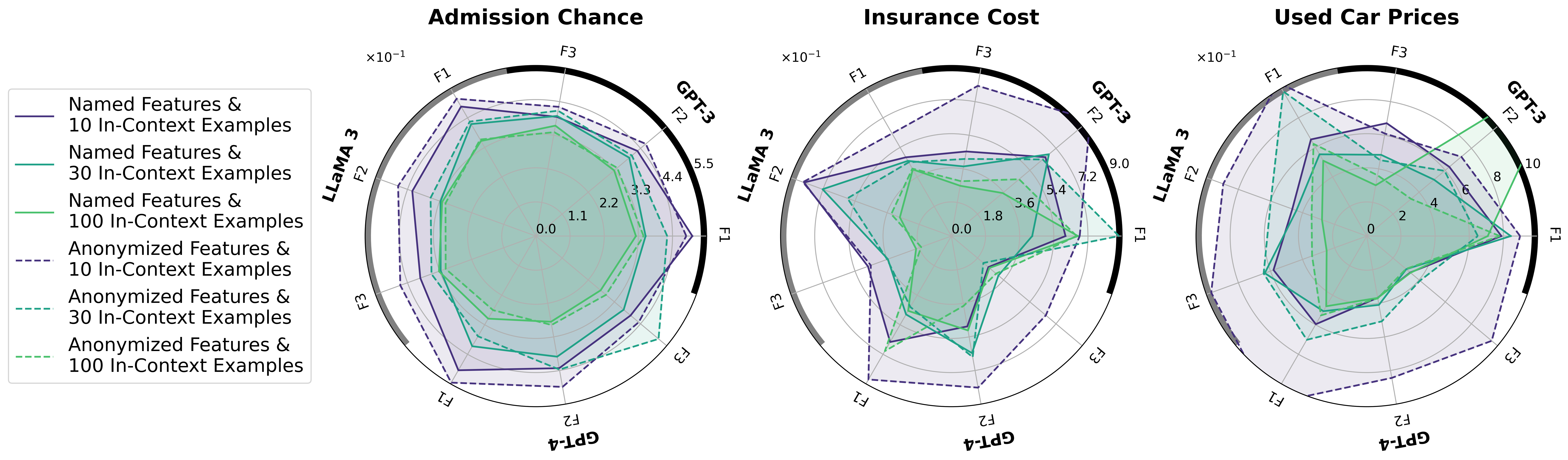}
    \caption{Figure 4 of the main paper with 1-$R^2$ instead of MSE. The figure shows the comparison of the number of in-context examples using Named Features (straight lines) and Anonymized Features (dashed lines) prompt configurations. F1, F2 and F3 represents using 1st (F1), then 1st and 2nd (F2), and all three feature (F3).
} 
    \label{fig:whole_context_r2}
\end{figure*}

\begin{figure}[!h]
    \centering
    \includegraphics[width=0.45\textwidth]{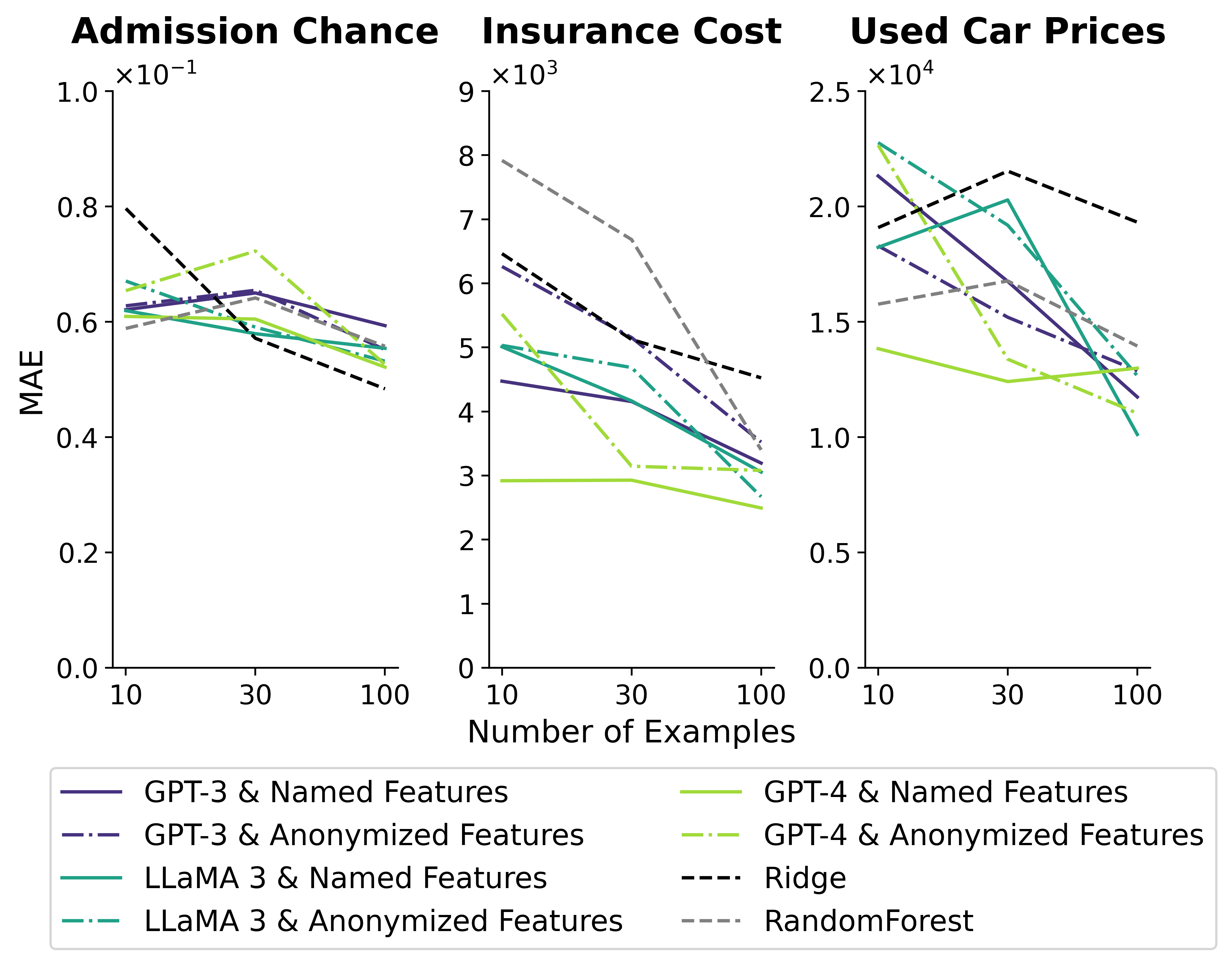}
    \caption{This figure shows Figure 5 of the main paper with MAE instead of MSE. The figure shows the performance of Named Features and Anonymized Features prompt configurations, Ridge, and RandomForest for 3 features based on the number of in-context examples.} 
    \label{fig:context_mae}
\end{figure}

\begin{figure}[!h]
    \centering
    \includegraphics[width=0.45\textwidth]{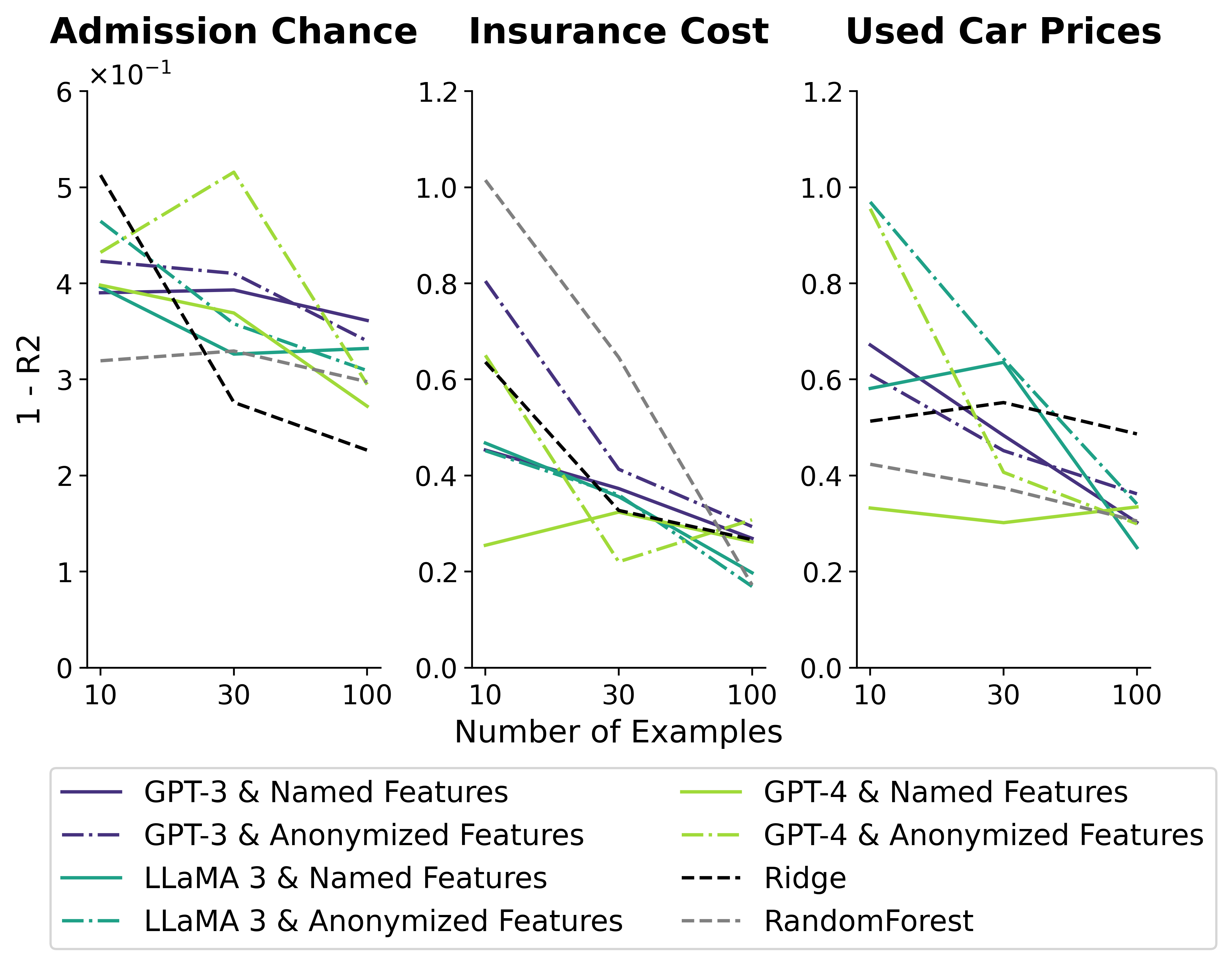}
    \caption{This figure shows Figure 5 of the main paper with 1-$R^2$ instead of MSE. The figure shows the performance of Named Features and Anonymized Features prompt configurations, Ridge, and RandomForest for 3 features based on the number of in-context examples.} 
    \label{fig:context_r2}
\end{figure}

\begin{figure}[!h]
    \centering
    \includegraphics[width=0.45\textwidth]{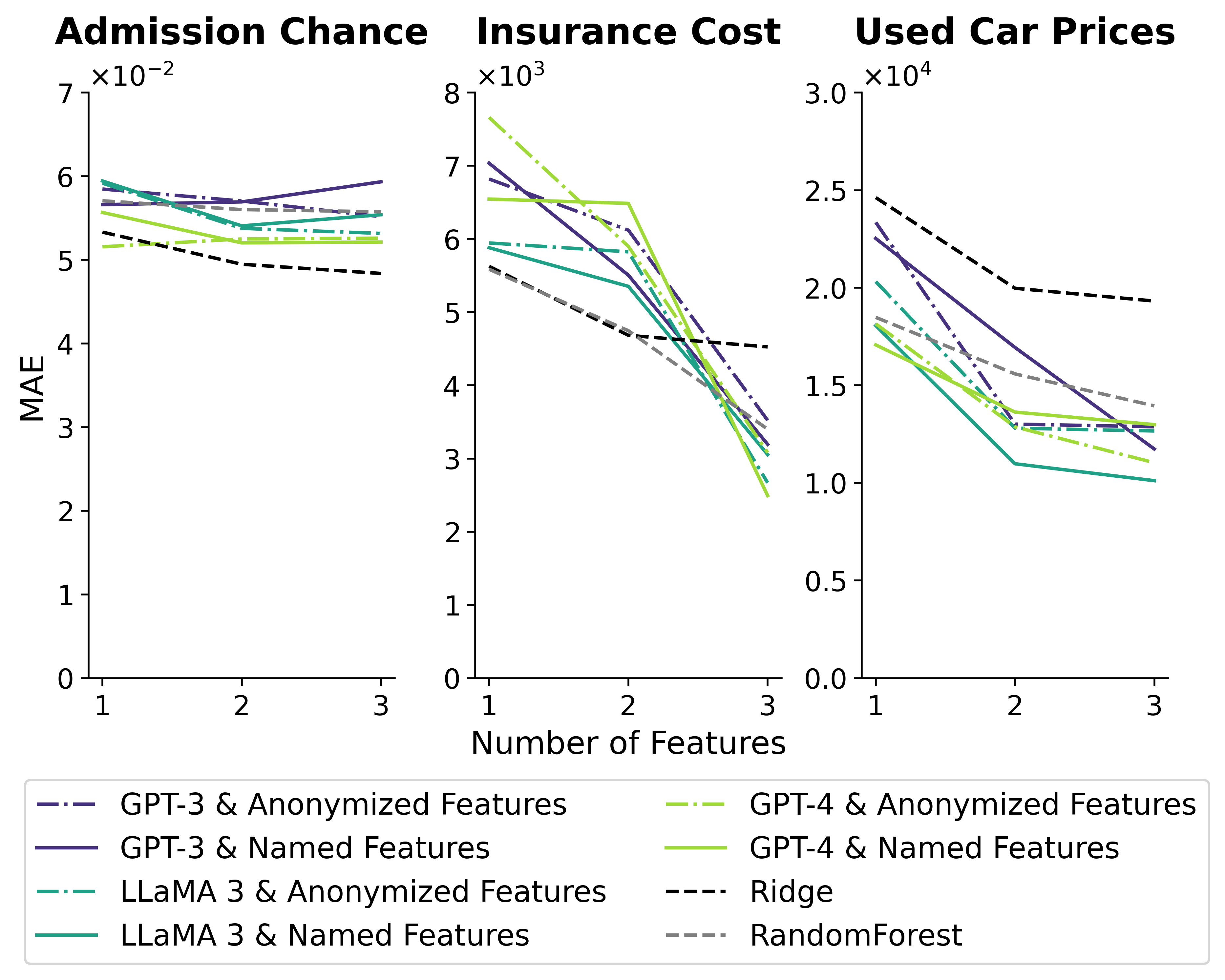}
    \caption{This figure shows Figure 6 of the main paper with MAE instead of MSE. The figure shows the performance of Anonymized Features prompt configuration, Ridge, and RandomForest for 100 in-context examples based on the number of features.} 
    \label{fig:features_mae}
\end{figure}

\begin{figure}[!h]
    \centering
    \includegraphics[width=0.45\textwidth]{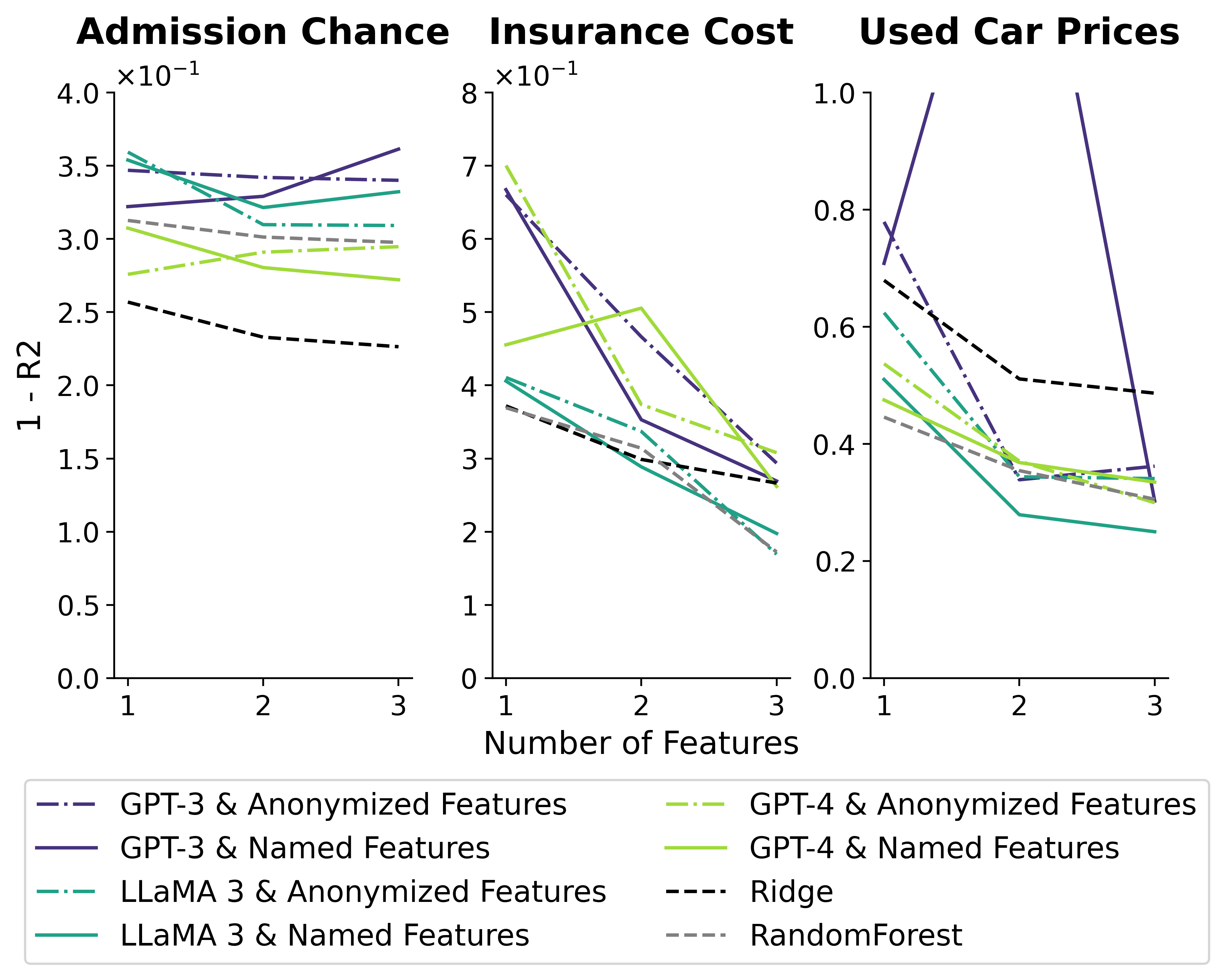}
    \caption{This figure shows Figure 6 of the main paper with 1-$R^2$ instead of MSE. The figure shows the performance of Anonymized Features prompt configuration, Ridge, and RandomForest for 100 in-context examples based on the number of features.} 
    \label{fig:features_r2}
\end{figure}

For Figures 2,3,4,5 and 6 of the main paper, the MAE results are shown in Figures \ref{fig:baseline_mae},\ref{fig:wholepicture_mae},\ref{fig:whole_context_mae},\ref{fig:context_mae} and \ref{fig:features_mae}, respectively. For Figures 2,3,4,5, and 6 of the main paper, the $1-R^2$ results are shown in Figures \ref{fig:baseline_r2},\ref{fig:wholepicture_r2},\ref{fig:whole_context_r2},\ref{fig:context_r2} and \ref{fig:features_r2}, respectively.

\section{Reasoning Configuration and Additional Diagrams}
\label{appendix:reasoning}
In this section, we include some diagrams that could have provided more insight and were removed from the main text for space. We also show examples of the reasoning configurations and diagrams related to that.

\subsection{Additional Diagrams of the Main Paper}

\begin{figure}[!h]
    \centering
    \includegraphics[width=0.45\textwidth]{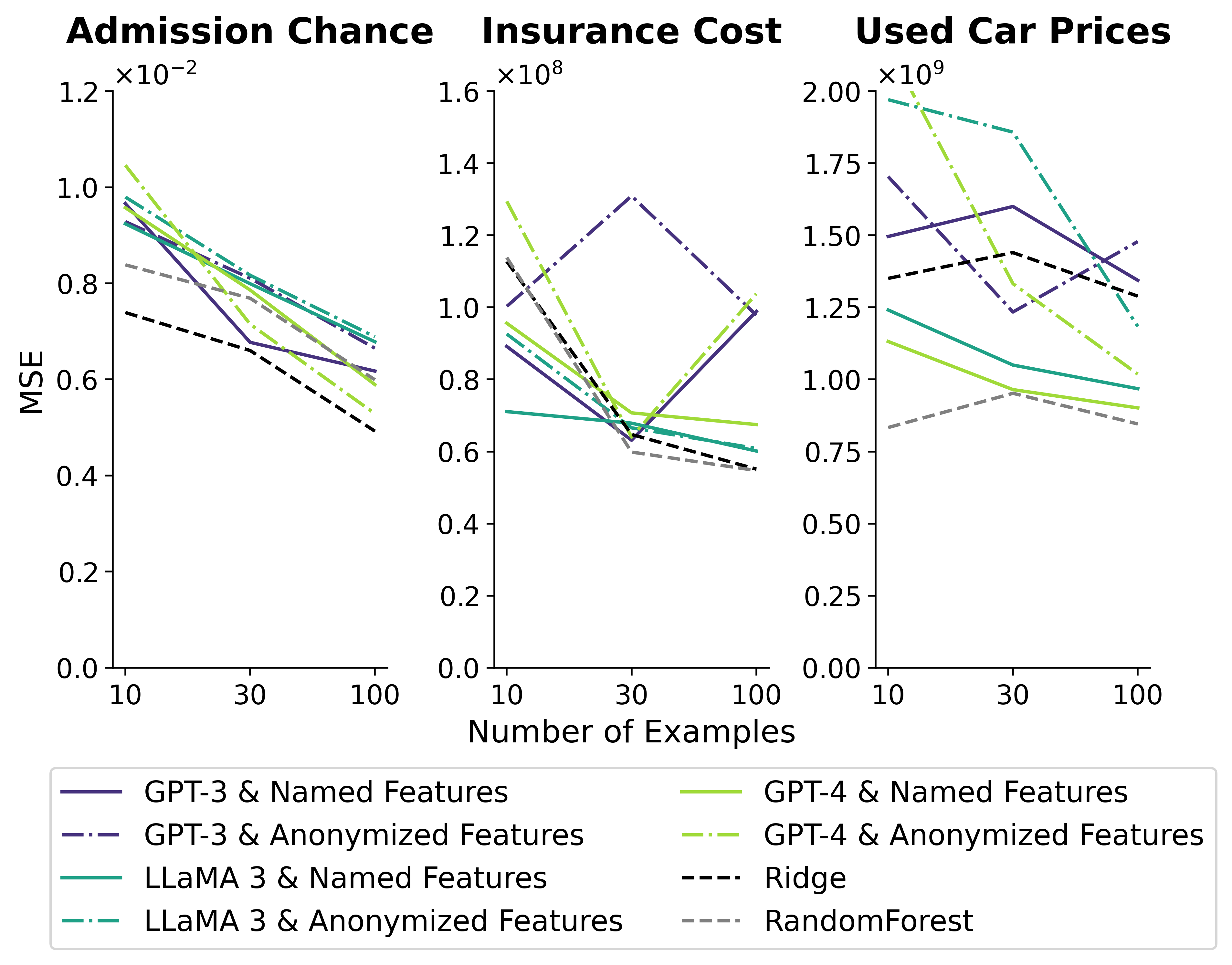}
    \caption{This figure shows Figure 5 of the main paper with 1 feature instead of 3. The figure shows the performance of Named Features and Anonymized Features prompt configurations, Ridge, and RandomForest based on the number of in-context examples.} 
    \label{fig:context_feature_1}
\end{figure}

\begin{figure}[!h]
    \centering
    \includegraphics[width=0.45\textwidth]{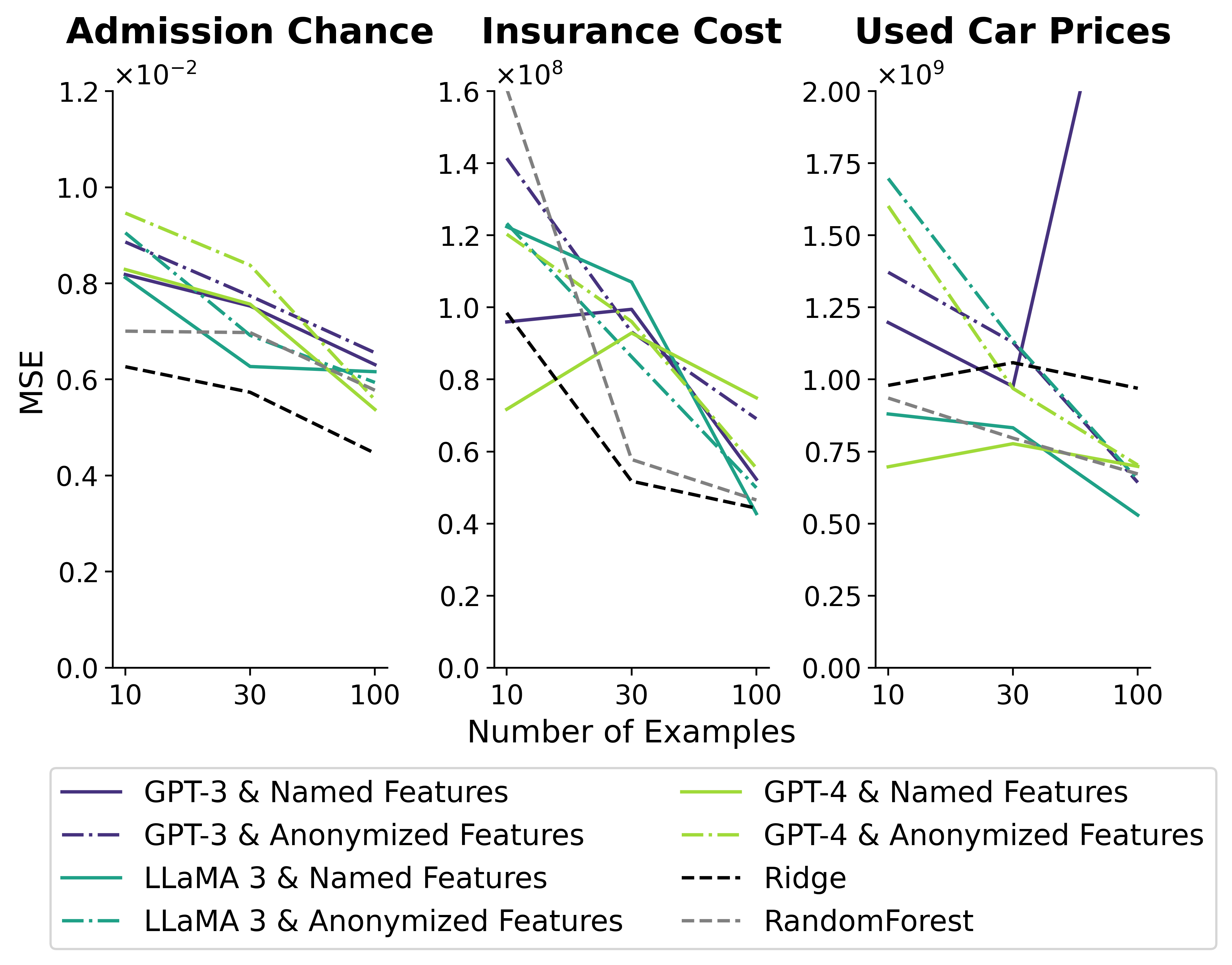}
    \caption{This figure shows Figure 5 of the main paper with 2 features instead of 3. The figure shows the performance of Named Features and Anonymized Features prompt configurations, Ridge, and RandomForest based on the number of in-context examples.} 
    \label{fig:context_feature_2}
\end{figure}

\begin{figure}[!h]
    \centering
    \includegraphics[width=0.45\textwidth]{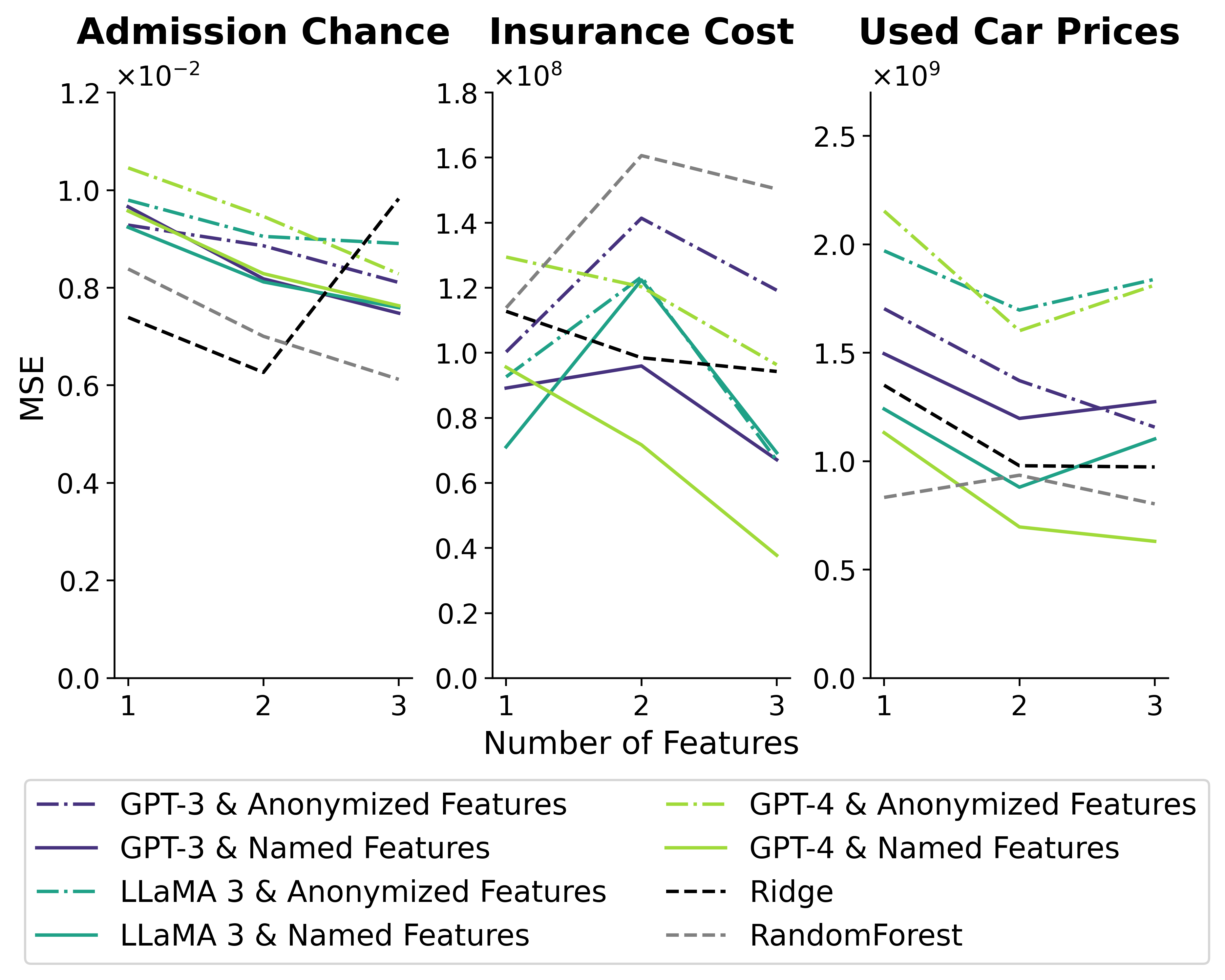}
    \caption{This figure shows Figure 6 of the main paper with 10 in-context examples instead of 100. The figure shows the performance of Anonymized Features prompt configuration, Ridge, and RandomForest based on the number of features.} 
    \label{fig:feature_context_10}
\end{figure}

\begin{figure}[!h]
    \centering
    \includegraphics[width=0.45\textwidth]{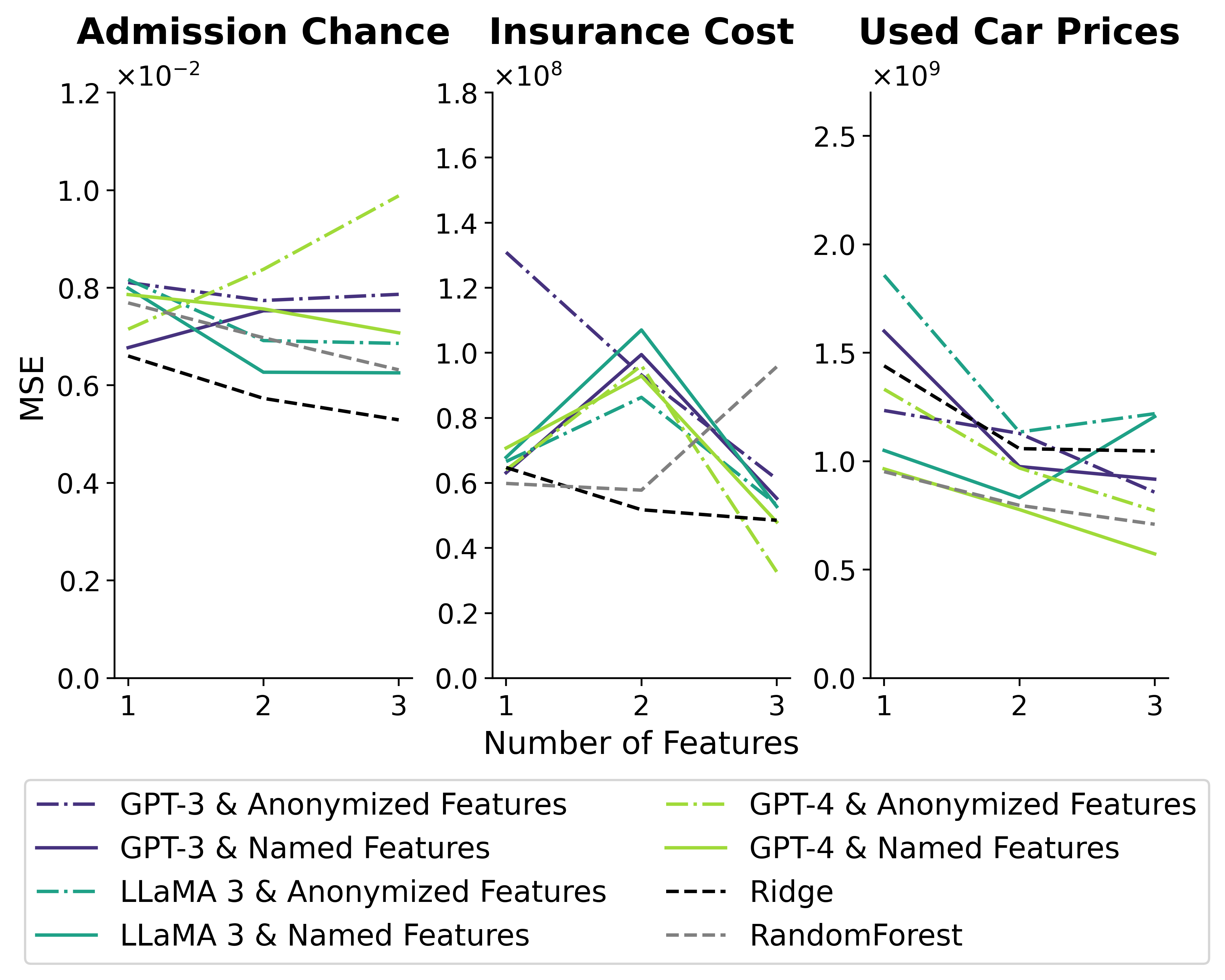}
    \caption{This figure shows Figure 6 of the main paper with 30 in-context examples instead of 100. The figure shows the performance of Anonymized Features prompt configuration, Ridge, and RandomForest based on the number of features.} 
    \label{fig:feature_context_30}
\end{figure}

Here, we included the variants of Figures 5 and 6 of the main paper that we omitted for space. Figure 5 with 1 and 2 features is shown in Figures \ref{fig:context_feature_1} and \ref{fig:context_feature_2}. Figure 6, with 10 and 30 in-context examples, is shown in Figures \ref{fig:feature_context_10} and \ref{fig:feature_context_30}.

\subsection{Reasoning Examples}

In this section, we show examples of reasoning for estimating an output for each dataset and LLM with 3 features. We will show the results of this setting in the next section.

\subsubsection{GPT-3 Reasoning over the Admission Chance Dataset:}
Based on the given information, the applicant has a high Cumulative GPA of 8.48, a decent GRE score of 311.0, and a good TOEFL score of 104.0. These scores are above average and indicate a strong academic background. 

Considering that the average Chance of Admission is 0.74 with a standard deviation of 0.14, and the applicant's strong academic profile, I would estimate that the Chance of Admission for this applicant is higher than the average. 

My final estimation is 0.80.

\subsubsection{LLaMA Reasoning over the Admission Chance Dataset:}
Based on the provided information, I will estimate the Chance of Admission. 

The given scores are all above average, indicating a strong applicant. The cumulative GPA of 8.48 is excellent, suggesting a high level of academic achievement. The GRE score of 311.0 is also above average, indicating strong verbal and quantitative skills. The TOEFL score of 104.0 is well above the average required for most graduate programs, demonstrating a high level of English proficiency.

Considering these strong scores, I would expect the Chance of Admission to be higher than the average of 0.74. Given the standard deviation of 0.14, I would estimate the Chance of Admission to be around one standard deviation above the mean.

My final estimation is 0.88.
\subsubsection{GPT-4 Reasoning over the Admission Chance Dataset:}

To estimate the Chance of Admission based on the provided information, we need to consider how the given data points (Cumulative GPA, GRE Score, TOEFL Score) might relate to the average Chance of Admission and its standard deviation. However, without specific information on how these factors correlate with admission chances, we'll have to make some general assumptions based on typical admissions criteria.

1. **Cumulative GPA**: An 8.48 GPA, assuming it's on a 10-point scale, is quite high. This would generally be seen as a strong positive in the admissions process, suggesting a higher than average chance of admission.

2. **GRE Score**: A score of 311 is somewhat above average (the average GRE scores for many programs tend to hover around the low to mid-300s). While not outstanding, it's competitive enough not to detract significantly from an application, especially with a strong GPA.

3. **TOEFL Score**: A TOEFL score of 104 is good, indicating a strong proficiency in English. For most universities, this score is above the minimum requirement and should be sufficient to not negatively impact the chance of admission.

Given the average Chance of Admission is 0.74 with a standard deviation of 0.14, and considering the strong GPA, adequate GRE, and TOEFL scores, it's reasonable to estimate the Chance of Admission to be above average but not exceptionally so due to the GRE score not being far above average. 

Taking into account the positive impact of a high GPA and good TOEFL score, balanced against the GRE score, a reasonable estimate might place the Chance of Admission slightly above the average but not by a large margin. Therefore, considering the given data and the average Chance of Admission, my final estimation is 0.78."

\subsubsection{GPT-3 Reasoning over the Insurance Cost Dataset:}
Based on the given information of being a smoker, having a BMI of 24.7, and being 43 years old, we can expect higher annual medical costs compared to non-smokers, individuals with lower BMIs, and younger individuals. Smokers tend to have higher medical costs due to the increased risk of various health conditions such as lung cancer, heart disease, and respiratory issues. Additionally, as individuals age, they are more likely to develop chronic health conditions that require ongoing medical care, which can also contribute to higher medical costs. A BMI of 24.7 is considered within the normal range, which may have a slightly lower impact on medical costs compared to individuals with higher BMIs.

Taking into account these factors, along with the average annual individual medical costs billed by health insurance in the USA of \$13,270.42 and a standard deviation of \$12,110.01, I estimate that the annual individual medical costs for a 43-year-old smoker with a BMI of 24.7 in the USA would be higher than the average due to the increased health risks associated with smoking and age. Therefore, my final estimation is \$20,000."
"Based on the provided information of being a non-smoker with a BMI of 25.8 and an age of 46, we can estimate that the individual is relatively healthy. Typically, younger individuals with a healthy BMI and who do not smoke tend to have lower medical costs compared to older individuals or those with unhealthy habits.

Considering the average annual individual medical costs billed by health insurance in the USA of \$13,270.42 with a standard deviation of \$12,110.01, we can assume that this individual's medical costs might be lower than the average due to their healthy profile. However, since medical costs can vary significantly based on individual health conditions and unforeseen circumstances, it is challenging to provide an accurate estimate without more specific information.

Taking into account the individual's healthy profile and the average costs in the USA, my final estimation is \$10,000."

\subsubsection{LLaMA Reasoning over the Insurance Cost Dataset:}

Based on the provided information, we can make an educated estimation of the annual individual medical costs billed by health insurance in the USA. 

We are given the average annual individual medical costs billed by health insurance in the USA, which is \$13,270.42, along with the standard deviation of \$12,110.01. This suggests that the distribution of medical costs is likely to be skewed to the right, with a long tail of high-cost individuals.

The additional information provided about the individual, such as being a smoker (smoker: 1.0), having a BMI of 24.7, and being 43 years old, can also influence the estimation. Smokers tend to have higher medical costs due to the increased risk of smoking-related illnesses. A BMI of 24.7 is considered overweight, which can also lead to higher medical costs. The age of 43 is relatively middle-aged, which may not have a significant impact on medical costs compared to older or younger individuals.

Considering these factors, I would adjust the average annual individual medical costs upwards to account for the increased risk associated with smoking and being overweight. However, I would not adjust it too drastically, as the individual's age is not extremely high.

My final estimation is \$16,500."

\subsubsection{GPT-4 Reasoning over the Insurance Cost Dataset:}
To estimate the annual individual medical costs billed by health insurance in the USA for a 43-year-old smoker with a BMI of 24.7, we need to consider how the given factors (smoking status, BMI, and age) typically influence medical costs, based on the provided average and standard deviation.

1. **Smoking Status**: Being a smoker generally increases medical costs significantly due to the higher risk of diseases associated with smoking (e.g., lung cancer, heart disease).
2. **BMI**: A BMI of 24.7 is within the normal weight range (18.5-24.9), which might not significantly increase medical costs compared to those with higher BMI values.
3. **Age**: At 43 years old, an individual might start to experience more health issues than a younger person, but not as many as someone in a more advanced age group. This could mean slightly higher than average costs due to age but not the highest possible.

Given the average annual medical costs of \$13,270.42 and considering the significant impact of smoking, the effect of age, and the relatively neutral effect of a normal BMI, the costs for this individual would likely be higher than the average but not excessively so due to the mitigating factor of a healthy BMI.

Considering the standard deviation of \$12,110.01, which indicates a wide range of costs, and the fact that smoking is a major cost driver, it's reasonable to estimate that the costs for this individual would be above average but not at the extreme end of the spectrum.

Given these considerations, a reasonable estimate for the annual individual medical costs for this person, factoring in the significant increase due to smoking but moderated by age and a healthy BMI, would be somewhat above the average but not by the full extent of one standard deviation.

My final estimation is \$20,000."

\subsubsection{GPT-3 Reasoning over the Used Car Prices Dataset:}

To estimate the price of a used car that can be either a Toyota or Maserati in 2019, we can consider the following factors:

1. City Fuel Economy: The given city fuel economy of 14.0 suggests that the car may not be very fuel-efficient, which could potentially lower its price.

2. Mileage: The mileage of 0.0 indicates that the car is essentially brand new or has very low mileage, which generally tends to increase the price of a used car.

3. Is Toyota: The fact that the car is not a Toyota implies that it could be a Maserati, which is a luxury brand known for higher price tags compared to Toyota vehicles.

Considering these factors, we can infer that the car is likely a Maserati due to its non-Toyota status, which would generally have a higher price tag. Given that the average price of a used car that can be either a Toyota or Maserati in 2019 is \$50,014.51 with a standard deviation of \$42,279.49, and considering the low mileage and potential luxury status of the car, my final estimation for the price of this used car would be higher than the average, so I estimate it to be \$60,000."

\subsubsection{LLaMA Reasoning over the Used Car Prices Dataset:}

Based on the given information, we can see that the average price of a used car that can be either a Toyota or Maserati in 2019 is 50014.51 with a standard deviation of 42279.49. Since the mileage is 0.0, it's likely that the car is relatively new or has been well-maintained, which could increase its value. The city fuel economy is 14.0, which is a moderate value, but it's not a key factor in determining the price of a used car.

The most important factor is the brand, and since Is Toyota is 0.0, it means the car is a Maserati, which is a luxury brand known for its high-performance and premium vehicles. Maserati cars tend to hold their value well, and their prices are generally higher than those of Toyota cars.

Considering these factors, I would estimate the price of this used car to be higher than the average price. My final estimation is 65000."

\subsubsection{GPT-4 Reasoning over the Used Car Prices Dataset:}

Given the information, we have a car with a city fuel economy of 14.0, mileage of 0.0, and it is not a Toyota (implying it could be a Maserati since the options are either Toyota or Maserati). Maseratis are luxury vehicles and typically have a higher price point compared to Toyotas. The average price of a used car from the given data is \$50,014.51 with a standard deviation of \$42,279.49. 

Considering the car in question is a Maserati (as indicated by ""Is Toyota: 0.0""), and it has 0.0 mileage, which suggests it's either new or nearly new, we can expect the price to be higher than the average of the mixed pool of Toyota and Maserati used cars. The low city fuel economy is characteristic of high-performance or luxury vehicles, which aligns with it being a Maserati. 

Given these factors, and knowing that Maseratis are premium vehicles, the price would likely be above the average. However, without knowing the exact model or year (beyond it being a used car as of 2019), we can't pinpoint the exact price but can infer it would be higher due to the brand and the newness indicated by zero mileage.

Considering the standard deviation and the average, and the fact that this is a luxury vehicle likely on the higher end of the price spectrum, a reasonable estimation, taking into account the luxury brand premium and the new condition, would be one standard deviation above the mean. Thus, my final estimation is \$92,294."

\subsection{Reasoning Configuration Diagrams}
\begin{figure*}[h]
    \centering
    \includegraphics[width=0.8\textwidth]{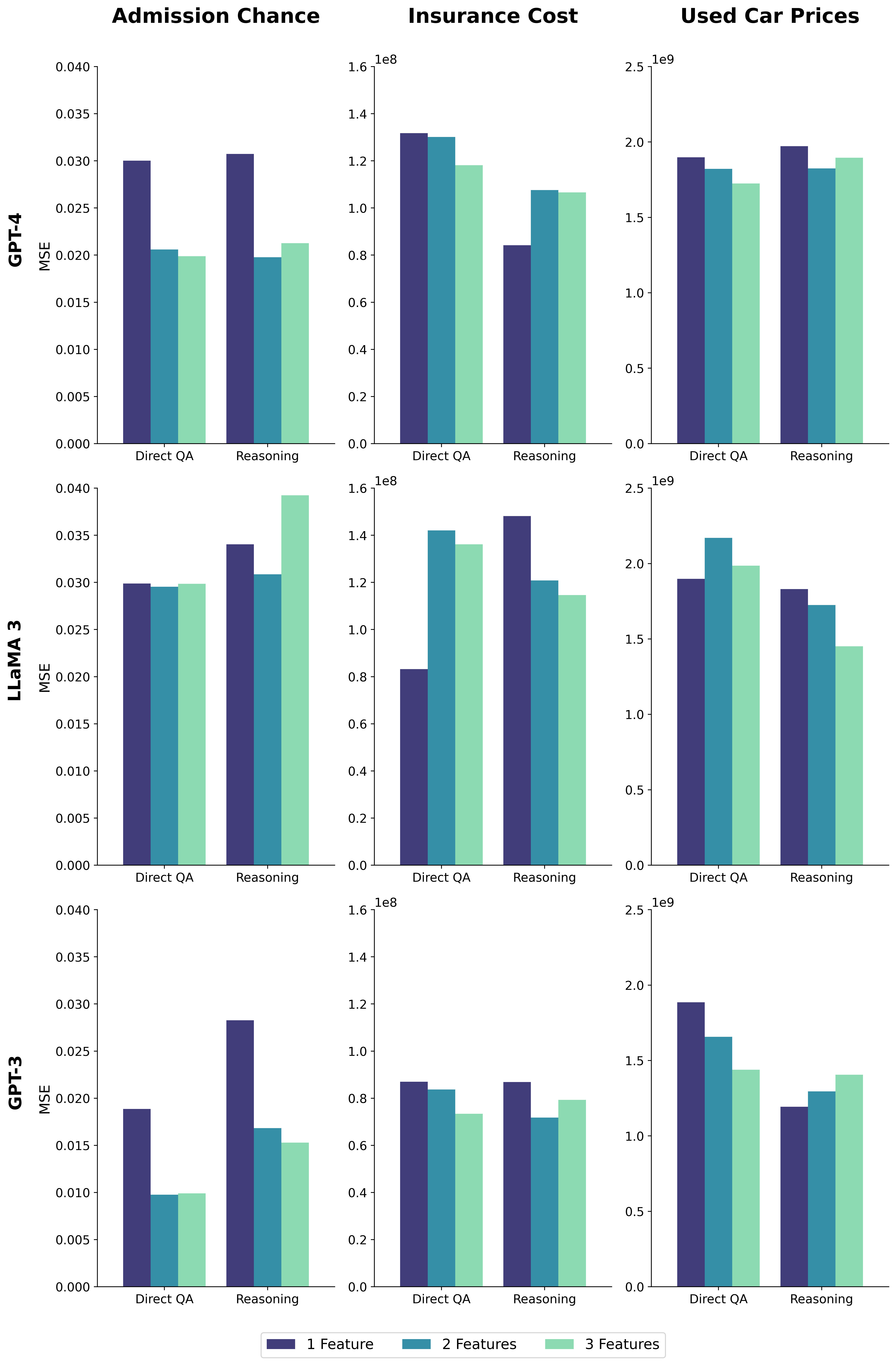}
    \caption{Comparison between Direct QA and Reasoning prompt configurations between dataset, LLMs and the number of features using MSE metric.
} 
    \label{fig:reasoning_diagram}
\end{figure*}

\begin{figure*}[h]
    \centering
    \includegraphics[width=0.8\textwidth]{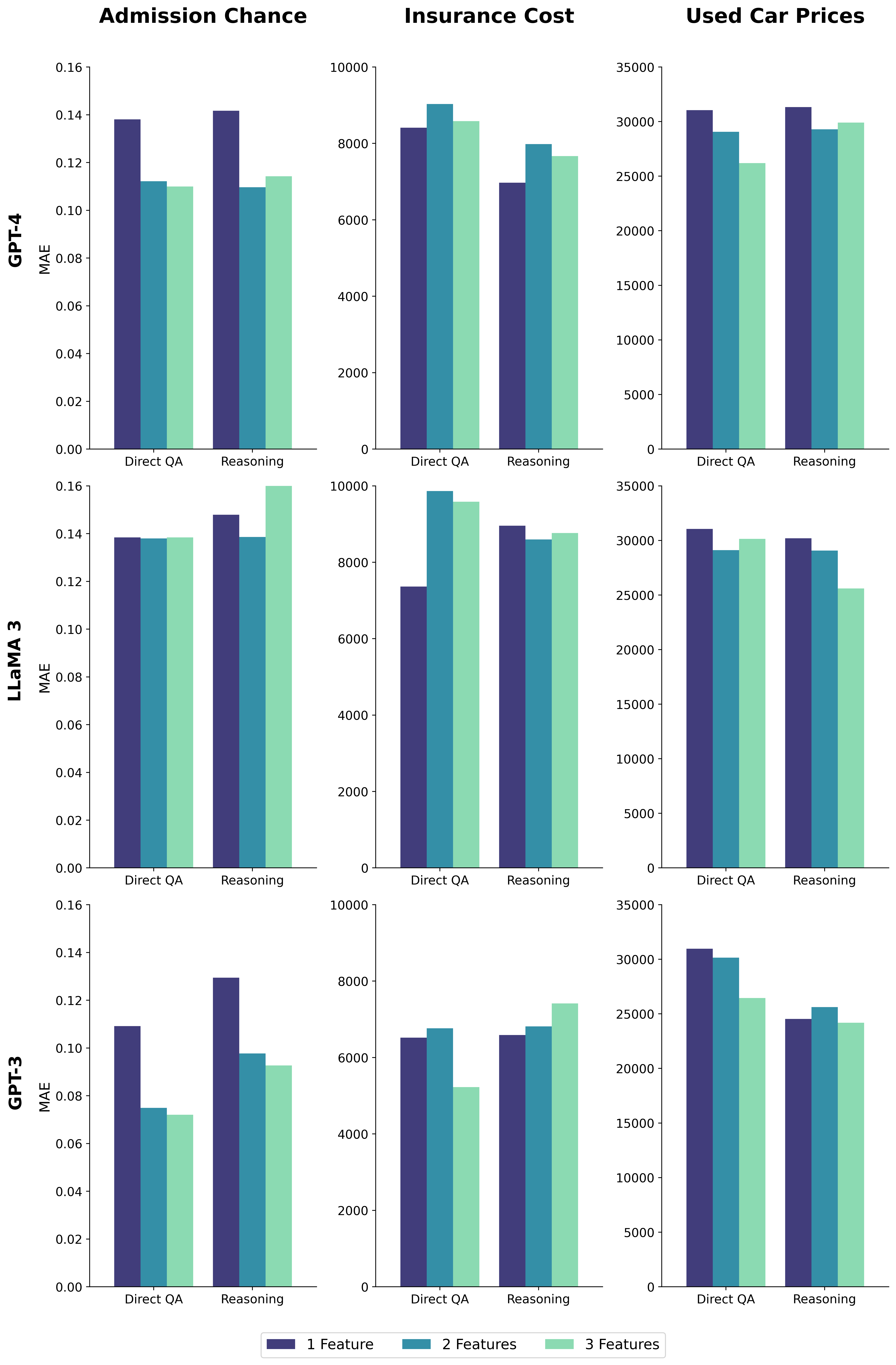}
    \caption{Comparison between Direct QA and Reasoning prompt configurations between dataset, LLMs and the number of features using MAE metric.
} 
    \label{fig:reasoning_diagram_MAE}
\end{figure*}

\begin{figure*}[h]
    \centering
    \includegraphics[width=0.8\textwidth]{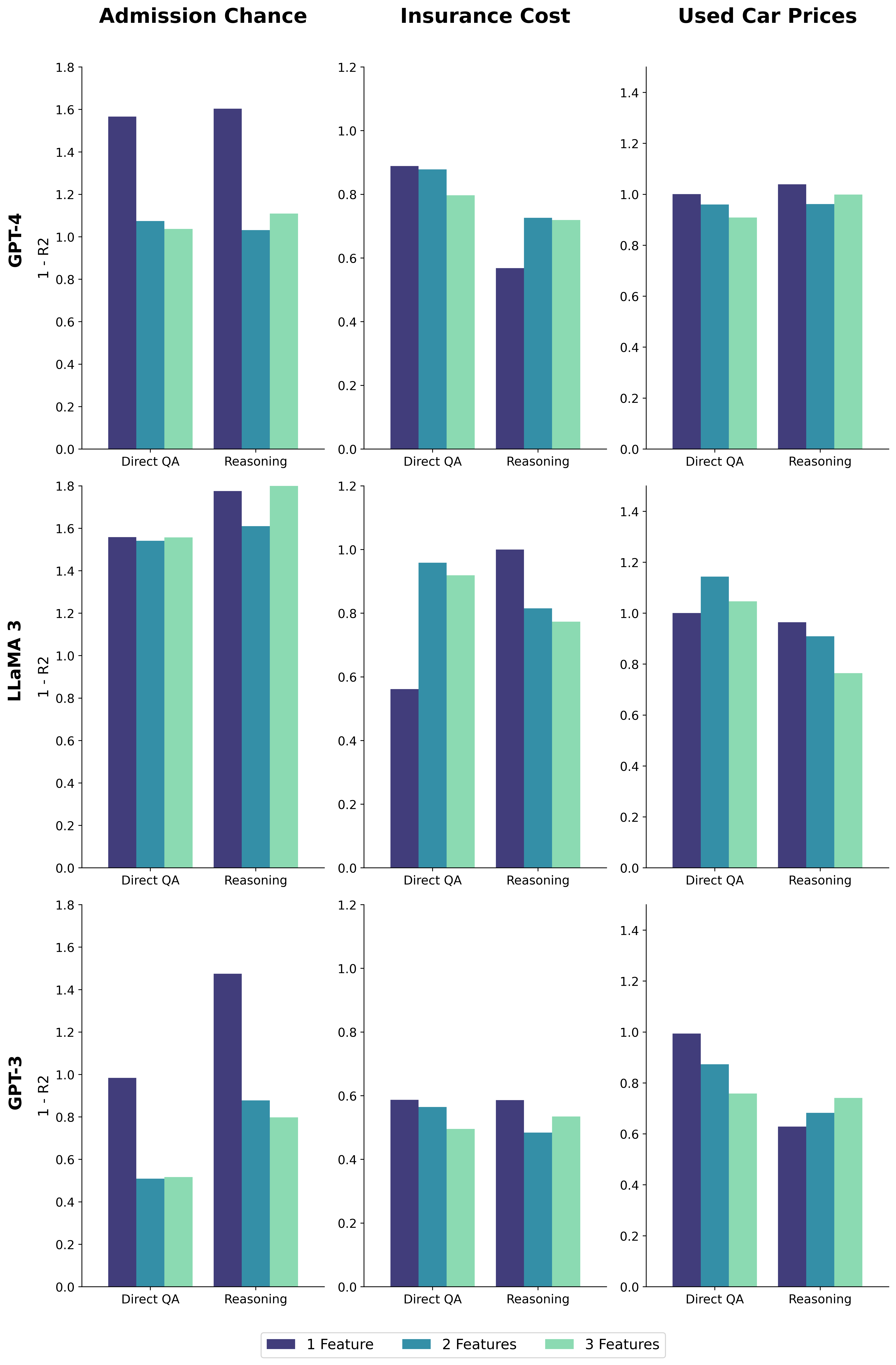}
    \caption{Comparison between Direct QA and Reasoning prompt configurations between dataset, LLMs and the number of features using $1-R^2$ metric.
} 
    \label{fig:reasoning_diagram_R2}
\end{figure*}

The diagrams that compare the Direct QA prompt configuration to the reasoning configuration are shown in Figures \ref{fig:reasoning_diagram}, \ref{fig:reasoning_diagram_MAE} and \ref{fig:reasoning_diagram_R2} for metrics MSE, MAE and $1-R^2$, respectively. As mentioned in the text of the main paper, there is no rhyme or reason for the behavior of the LLMs in this regard.

\end{document}